\documentclass{article}

\usepackage{microtype}
\usepackage{graphicx}
\usepackage{subcaption}
\usepackage{booktabs} 
\usepackage{arydshln}
\usepackage{multirow}
\usepackage{diagbox}
\usepackage{colortbl}
\usepackage[table]{xcolor}
\usepackage{array}       
\usepackage{rotating}    
\usepackage{siunitx}
\usepackage{adjustbox}
\usepackage{hyperref}

\usepackage[preprint]{icml2026}

\usepackage{amsmath}
\usepackage{amssymb}
\usepackage{mathtools}
\usepackage{amsthm}

\usepackage[capitalize,noabbrev]{cleveref}
\usepackage[most]{tcolorbox} 
\usepackage{xcolor}          
\usepackage{tikz}            
\definecolor{colorhead}{HTML}{e2ecda}

\definecolor{customgreen}{RGB}{46, 163, 67} 

\theoremstyle{plain}

\theoremstyle{definition}

\theoremstyle{remark}

\usepackage[textsize=tiny]{todonotes}

\begin{document}

\twocolumn[
  \icmltitle{VEQ: Modality-Adaptive Quantization for MoE Vision-Language Models}



  \icmlsetsymbol{equal}{*}

  \begin{icmlauthorlist}
    \icmlauthor{Guangshuo Qin}{equal,sjtu}
    \icmlauthor{Zhiteng Li}{equal,sjtu}
    \icmlauthor{Zheng Chen}{sjtu}
    \icmlauthor{Weihang Zhang}{tsinghua}
    \icmlauthor{Linghe Kong}{sjtu}
    \icmlauthor{Yulun Zhang~\textsuperscript{\textdagger}}{sjtu}
  \end{icmlauthorlist}

  \icmlaffiliation{sjtu}{Shanghai Jiao Tong University}
  \icmlaffiliation{tsinghua}{Tsinghua University}

  \icmlcorrespondingauthor{Yulun Zhang}{yulun100@gmail.com}

  \icmlkeywords{Machine Learning, ICML}
  \vskip 0.3in
]



\printAffiliationsAndNotice{\icmlEqualContribution}

\begin{abstract}
Mixture-of-Experts(MoE) Vision-Language Models (VLMs) offer remarkable performance but incur prohibitive memory and computational costs, making compression essential. Post-Training Quantization (PTQ) is an effective training-free technique to address the massive memory and computation overhead. Existing quantization paradigms fall short as they are oblivious to two critical forms of heterogeneity: the inherent discrepancy between vision and language tokens, and the non-uniform contribution of different experts. To bridge this gap, we propose Visual Expert Quantization (VEQ), a dual-aware quantization framework designed to simultaneously accommodate cross-modal differences and heterogeneity between experts. Specifically, VEQ incorporates 1)\textbf{Modality-expert-aware Quantization}, which utilizes expert activation frequency to prioritize error minimization for pivotal experts, and 2)\textbf{Modality-affinity-aware Quantization}, which constructs an enhanced Hessian matrix by integrating token-expert affinity with modality information to guide the calibration process. Extensive experiments across diverse benchmarks verify that VEQ consistently outperforms state-of-the-art baselines. Specifically, under the W3A16 configuration, our method achieves significant average accuracy gains of 2.04\% on Kimi-VL and 3.09\% on Qwen3-VL compared to the previous SOTA quantization methods, demonstrating superior robustness across various multimodal tasks. Our code will be available at \hyperlink{https://github.com/guangshuoqin/VEQ}{https://github.com/guangshuoqin/VEQ}.
\end{abstract}

\setlength{\abovedisplayskip}{2pt}
\setlength{\belowdisplayskip}{2pt}

\section{Introduction}
In recent years, Vision-Language Models (VLMs) have demonstrated unprecedented proficiency across a broad spectrum of multimodal tasks, ranging from fundamental visual question answering and image captioning to complex visual reasoning \cite{radford_learning_2021,alayrac_flamingo_2022,li_blip-2_2023}. By seamlessly aligning visual perception with linguistic semantics, these models effectively bridge the cross-modal divide, empowering intelligent agents to perceive, reason, and interact with the physical environment with human-level versatility \cite{lu_vilbert_2019,openai_gpt-4o_2024}. With the increasing demand for robust multimodal understanding, VLMs have become increasingly important for practical applications.

\begin{figure}[t]
    \centering
    
    \includegraphics[width=1.0\linewidth]{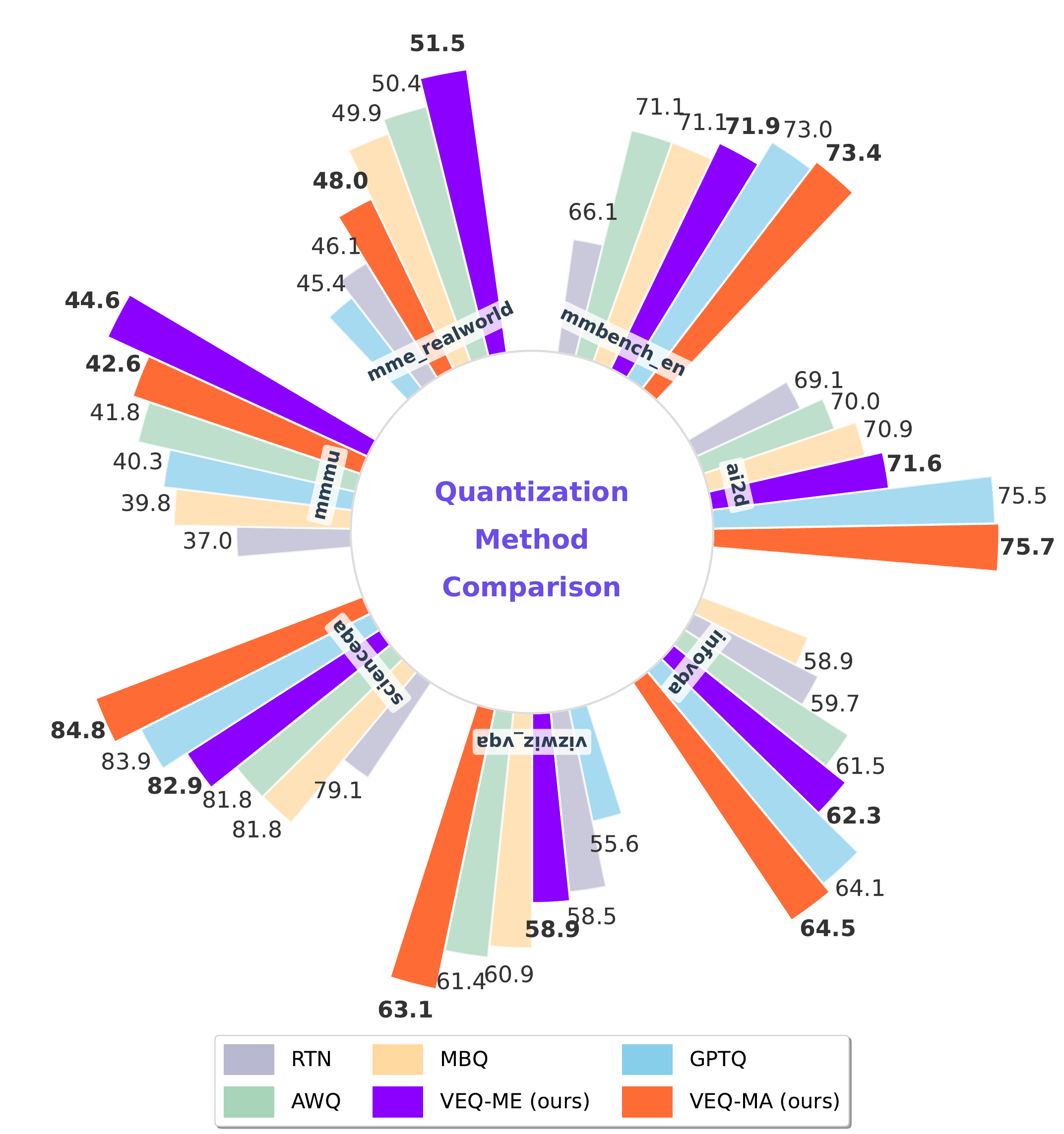}
    \vspace{-6mm}
    \caption{Zero-shot performance of Kimi-VL-Instruct under 3-bit weight quantization (W3A16). Our methods consistently outperform established baselines, demonstrating superior robustness.}
    \label{fig:quantization_comparison}
    \vspace{-8mm}
\end{figure}
To further scale up model capacity while maintaining computational efficiency, the Mixture-of-Experts (MoE) architecture has been widely adopted in state-of-the-art VLMs. Unlike dense models that activate all parameters for every token, MoE models utilize a sparse routing mechanism to activate only a subset of experts, effectively reducing inference costs while preserving a vast parameter space \cite{fedus_switch_2022}. Prominent open-source VLMs, such as DeepSeek-VL2 \cite{deepseek-vl2}, Kimi-VL \cite{team2025kimi}, Qwen3-VL \cite{Qwen3-VL}, and ERNIE-4.5-VL \cite{ernie2025technicalreport}, have successfully leveraged this architecture to achieve superior performance with manageable resource consumption.
\begin{figure}[t]
\vspace{-1.5mm}
    \centering
    \begin{subfigure}[b]{0.48\columnwidth}
        \centering
        \includegraphics[width=\linewidth]{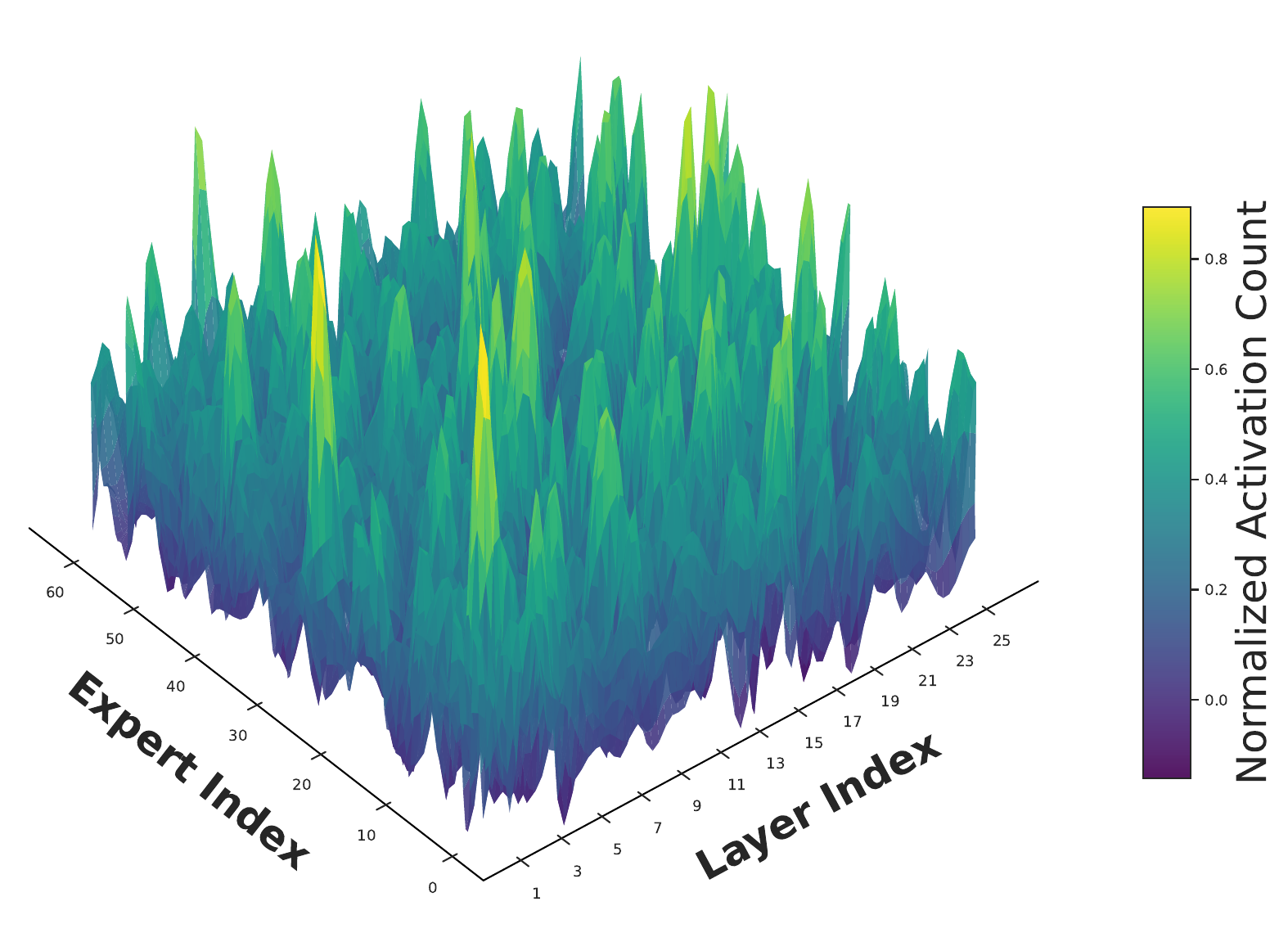}
        \caption{Text token.}
        \label{fig:text_expert}
    \end{subfigure}
    \begin{subfigure}[b]{0.48\columnwidth}
        \centering
        \includegraphics[width=\linewidth]{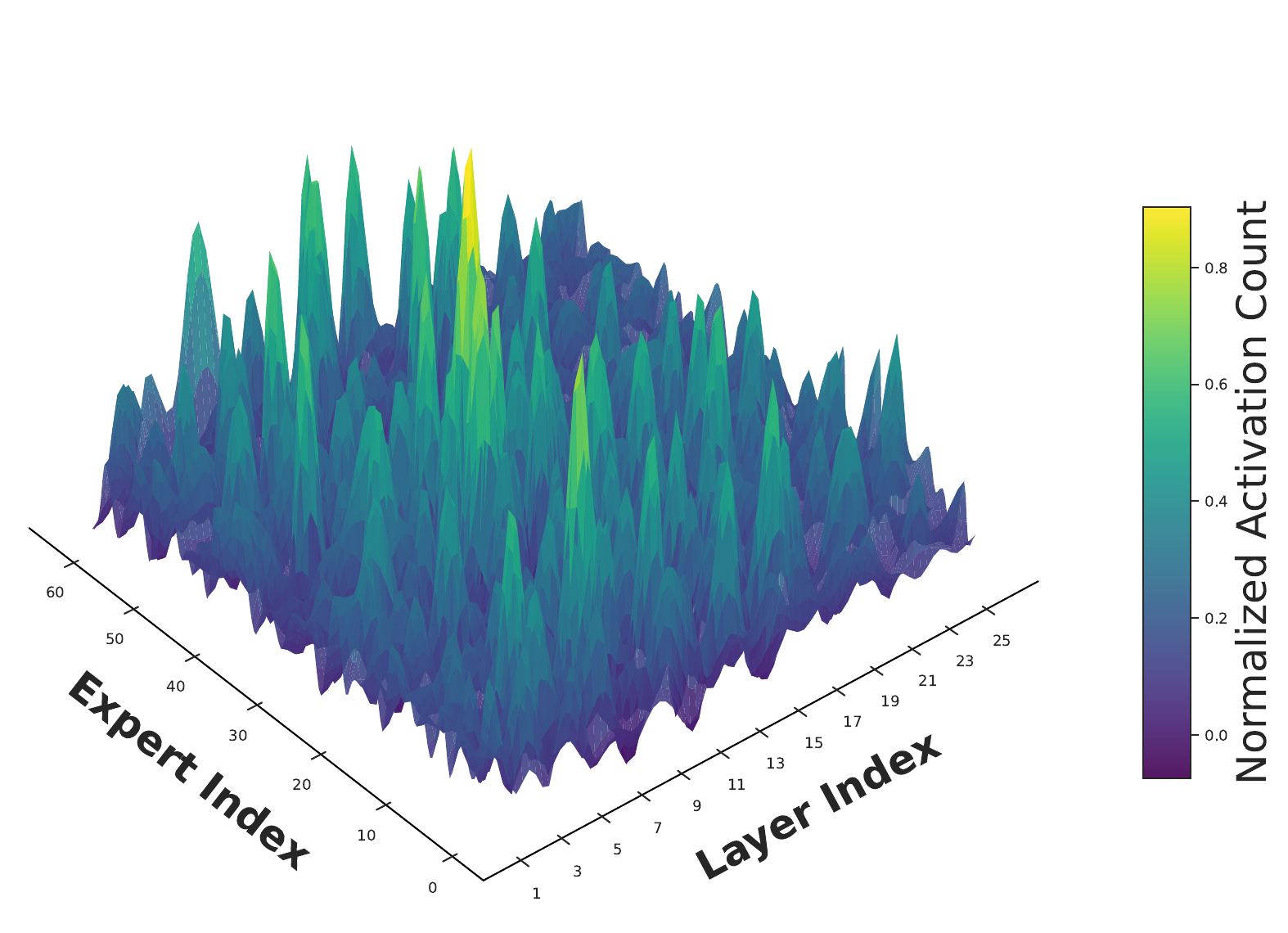}
        \caption{Vision token.}
        \label{fig:vision_expert}
    \end{subfigure}
    \vspace{-2mm}
    \caption{Comparative analysis of activation characteristics across different modalities. Peaks represent high activation frequency.}
    \label{fig:motivation_activation_surfaces}
    \vspace{-7mm}
\end{figure}
Despite their efficiency advantages over dense counterparts, MoE VLMs still incur significant memory footprints and latency during inference, necessitating effective model compression techniques. Post-Training Quantization (PTQ) has emerged as a practical solution. Mainstream weight-only quantization methods, such as AWQ \cite{lin_awq_2024} and GPTQ \cite{frantar_gptq_2023}, primarily focus on minimizing weight reconstruction error to preserve model performance. Meanwhile, activation-aware methods like SmoothQuant \cite{xiao_smoothquant_2024} and SpinQuant \cite{liu_spinquant_2025} address the challenge of activation outliers by smoothing or rotating the feature space. More recently, MBQ \cite{li2025mbq} introduced a novel perspective by exploring the sensitivity of multimodal tokens to further minimize quantization error.

However, directly applying these conventional quantization methods to MoE VLMs proves suboptimal. Existing approaches typically treat the model as a monolithic dense structure, thereby neglecting the inherent structural sparsity of MoE architectures. Specifically, they fail to account for the varying importance of different experts. As illustrated in Figure \ref{fig:motivation_activation_surfaces}, a small subset of hot experts is accessed more frequently and dominates the output, while other experts remain dormant \cite{fedus_switch_2022,chi_representation_2022}. Furthermore, current methods often neglect the distinct statistical distribution heterogeneity between vision and text tokens \cite{liang_mind_2022}. Applying a unified quantization strategy across these modalities leads to significant performance deterioration, as the sensitivity to quantization noise varies drastically between continuous visual embeddings and discrete textual representations.

To address these challenges, we propose Visual Expert Quantization (\textbf{VEQ}), a novel framework tailored for MoE VLMs. 
We conduct comprehensive evaluations on leading MoE architectures, including Kimi-VL \cite{team2025kimi} and Qwen3-VL \cite{Qwen3-VL}, across challenging multimodal benchmarks such as MMMU \cite{yue_mmmu_2024}, MME-RealWorld \cite{zhang2024mme}, MMBench \cite{liu2024mmbench}, and InfoVQA \cite{mathew_infographicvqa_2021}. 
Empirical results demonstrate that VEQ consistently outperforms established baselines (e.g., AWQ \cite{lin_awq_2024}, GPTQ \cite{frantar_gptq_2023}, MBQ \cite{li2025mbq}), establishing a new state-of-the-art for quantized MoE VLMs. 
Notably, VEQ exhibits superior robustness in low-bit settings. 
To the best of our knowledge, this represents a pioneering effort in compressing large-scale MoE VLMs.

Our main contributions are summarized as follows:
\begin{itemize}
\vspace{-3mm}
    \item \textbf{Pioneering Framework for MoE VLMs:} To the best of our knowledge, this work represents the first attempt to simultaneously address the dual challenges of multimodal heterogeneity and the unique structural properties of Mixture-of-Experts (MoE) based architectures in the context of quantization. 
 \vspace{-2mm}   
    \item \textbf{Modality-Expert-Aware Quantization:} We propose a novel strategy that assigns importance scores to experts based on the routing frequency of visual and textual tokens. By explicitly modeling these routing patterns, we utilize the scores to effectively minimize quantization error in critical experts.
\vspace{-2mm}    
    \item \textbf{Modality-Affinity-Aware Quantization:} We introduce a refinement method that leverages router affinity logits and input token modalities to re-weight the Hessian matrix. This approach incorporates semantic affinity into the optimization process, further enhancing the precision of the quantized model.
\end{itemize}

\begin{figure*}[t]
    \centering
    \includegraphics[width=\textwidth]{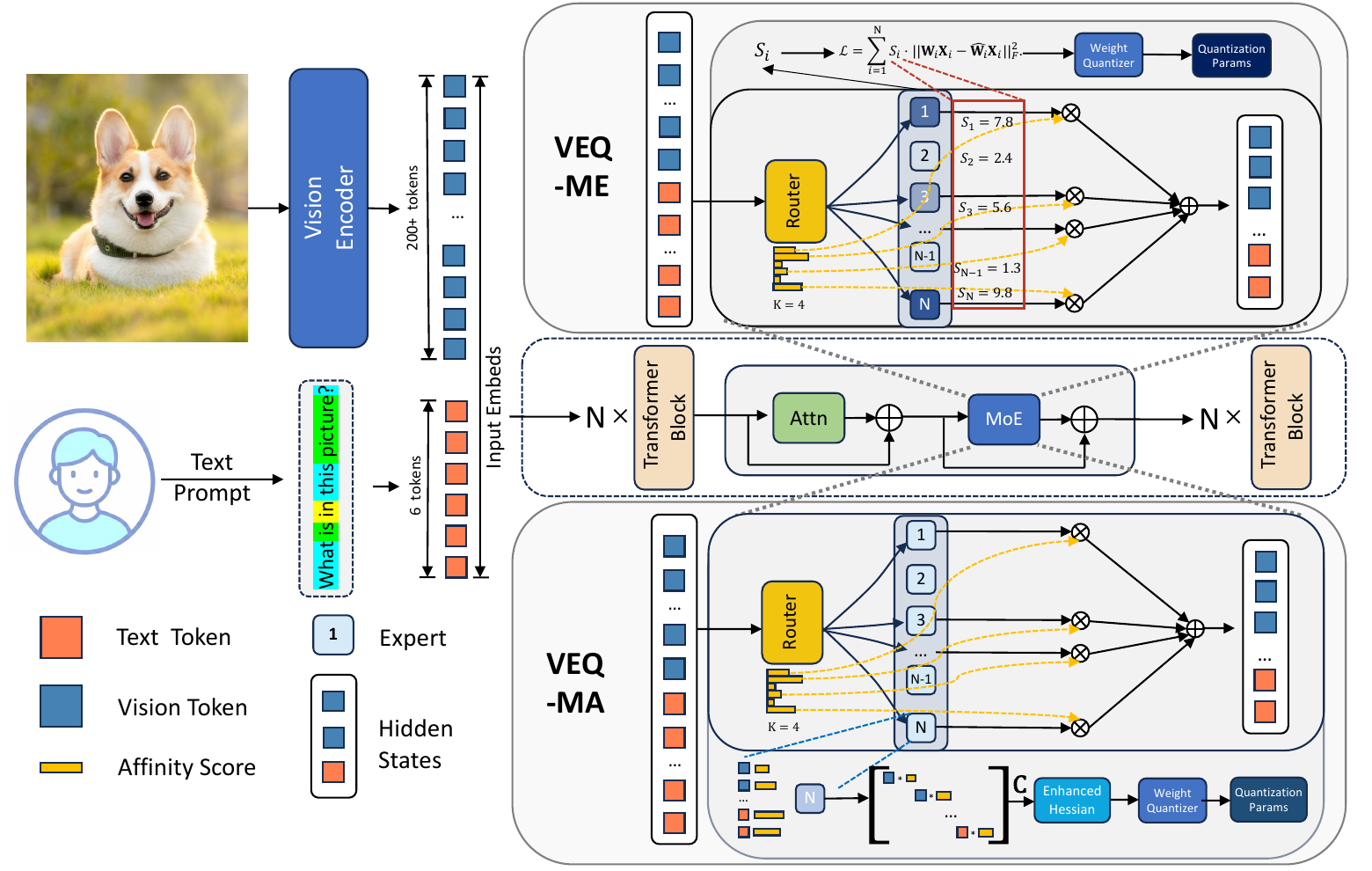}
    \vspace{-4mm}
    \caption{Overview of the proposed VEQ framework. Our method consists of two core components: (1) \textbf{VEQ-ME}, which dynamically assigns importance scores $S_i$ to experts based on their activation frequencies, thereby prioritizing error minimization for pivotal experts in the reconstruction loss; and (2) \textbf{VEQ-MA}, which constructs an enhanced Hessian matrix by integrating token-expert affinity scores and modality sensitivity, enabling the calibration process to adapt to the varying sensitivities of multi-modal tokens.}
    \label{fig:method_overview}
    \vspace{-5mm}
\end{figure*}
\vspace{-4mm}
\section{Related Work}
\vspace{-2mm}
\subsection{VLM Quantization} 
\vspace{-2mm}
Post-Training Quantization (PTQ) for Vision-Language Models (VLMs) remains a challenging and relatively underexplored area, primarily due to the distribution heterogeneity between vision and text modalities. VLMQ~\cite{xue2025vlmq} addresses visual token redundancy by proposing an importance-aware objective. It generates an enhanced Hessian matrix incorporating token-level importance factors via a lightweight block-wise backward pass. Q-VLM~\cite{wang2024q-vlm} utilizes activation entropy as a proxy to identify cross-layer dependencies for efficient block partitioning. It further optimizes the visual encoder to disentangle these dependencies, thereby reducing search overhead while maintaining accuracy. MBQ~\cite{li2025mbq} accounts for the distinct sensitivity levels of vision and language tokens by incorporating gradient-based sensitivity indicators into the calibration process, aiming to balance reconstruction loss across modalities. Bi-VLM~\cite{wang2025bi-vlm} implements a saliency-aware hybrid quantization algorithm that partitions weights non-uniformly based on Gaussian quantiles. This approach assigns higher precision to salient outliers while binarizing the remaining parameters. MQuant~\cite{yu2025mquant} introduces modality-specific static quantization and an attention-invariant switching mechanism to address distribution disparities. Additionally, it employs Rotation Magnitude Suppression (RMS) to mitigate outliers induced by online Hadamard transformations.

\subsection{MoE LLM Quantization} 
\vspace{-1.5mm}
PTQ also presents significant difficulties for Mixture-of-Experts (MoE)  architectures, stemming from intrinsic expert sparsity and the complex affinity between tokens and experts. Several recent studies have aimed to address these issues. MoEQuant~\cite{hu2025moequant} tackles inter- and intra-expert imbalances by employing an expert-balanced self-sampling method for calibration. It further utilizes an affinity-guided quantization strategy that weights errors according to token-expert correlations. MoQa~\cite{zheng2025moqa} implements an expert-level mixed-precision base quantization via multi-stage data-model distribution analysis, complemented by a channel-level dynamic adjustment mechanism to adapt to novel data distributions. MoQE~\cite{zhang2025moqe} leverages the MoE architecture for inference acceleration by treating multiple quantization variants of a single model as experts, using a lightweight router to dynamically assign input data to the optimal quantization expert. MxMoE~\cite{duanmu2025mxmoe} proposes an accuracy-performance co-design framework that allocates bit-widths at the granularity of linear blocks. This is achieved by analyzing parameter sensitivity and expert activation frequencies to optimize mixed-precision configurations for quantization.

\section{Method}\label{sec:method}
\vspace{-2mm}
In this section, we propose \textbf{Visual Expert Quantization (VEQ)}, a novel post-training quantization framework tailored for MoE VLMs. We begin by analyzing the intrinsic heterogeneity of MoE VLM in~\cref{modality experts heterogeneity analysis}. Building on these observations, we formulate a modality-aware expert importance metric in \cref{Modality-Expert-Aware Quantization}. Finally, we introduce an affinity-aware quantization algorithm to minimize reconstruction error in \cref{Modality-Affinity-Aware Quantization}.

\vspace{-2.5mm}
\subsection{Heterogeneity in MoE VLM Quantization}\label{modality experts heterogeneity analysis}
\vspace{-1.5mm}
\subsubsection{Modality Heterogeneity}
\vspace{-1.5mm}
Distinct modalities exhibit unique routing patterns and sensitivity levels within MoE layers. To quantify this heterogeneity, we utilize the gradient of the Supervised Fine-Tuning (SFT) loss as a metric to measure the error sensitivity of vision and text tokens. 

\begin{figure}[t]
    \centering
    \includegraphics[width=1.0\linewidth]{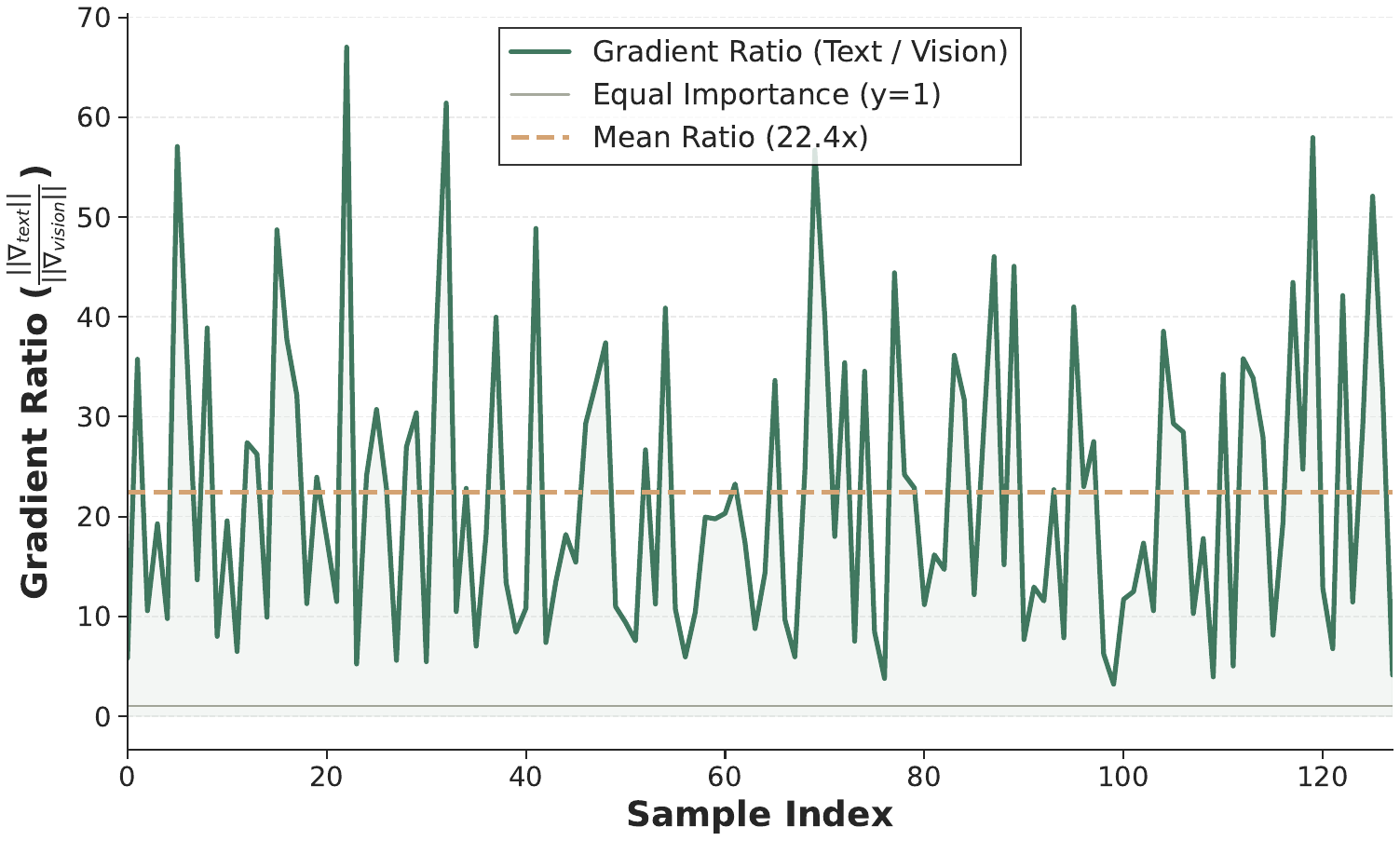}
    \vspace{-5mm}
    \caption{Analysis of gradient magnitude across 128 samples from the COCO \cite{lin2014coco} dataset. The text tokens exhibit significantly higher gradient norms compared to vision tokens, with an average ratio of 22.4.}
    \label{fig:gradient_ratio}
    \vspace{-7mm}
\end{figure}
\begin{figure}[t]
    \centering
    \includegraphics[width=1.0\linewidth]{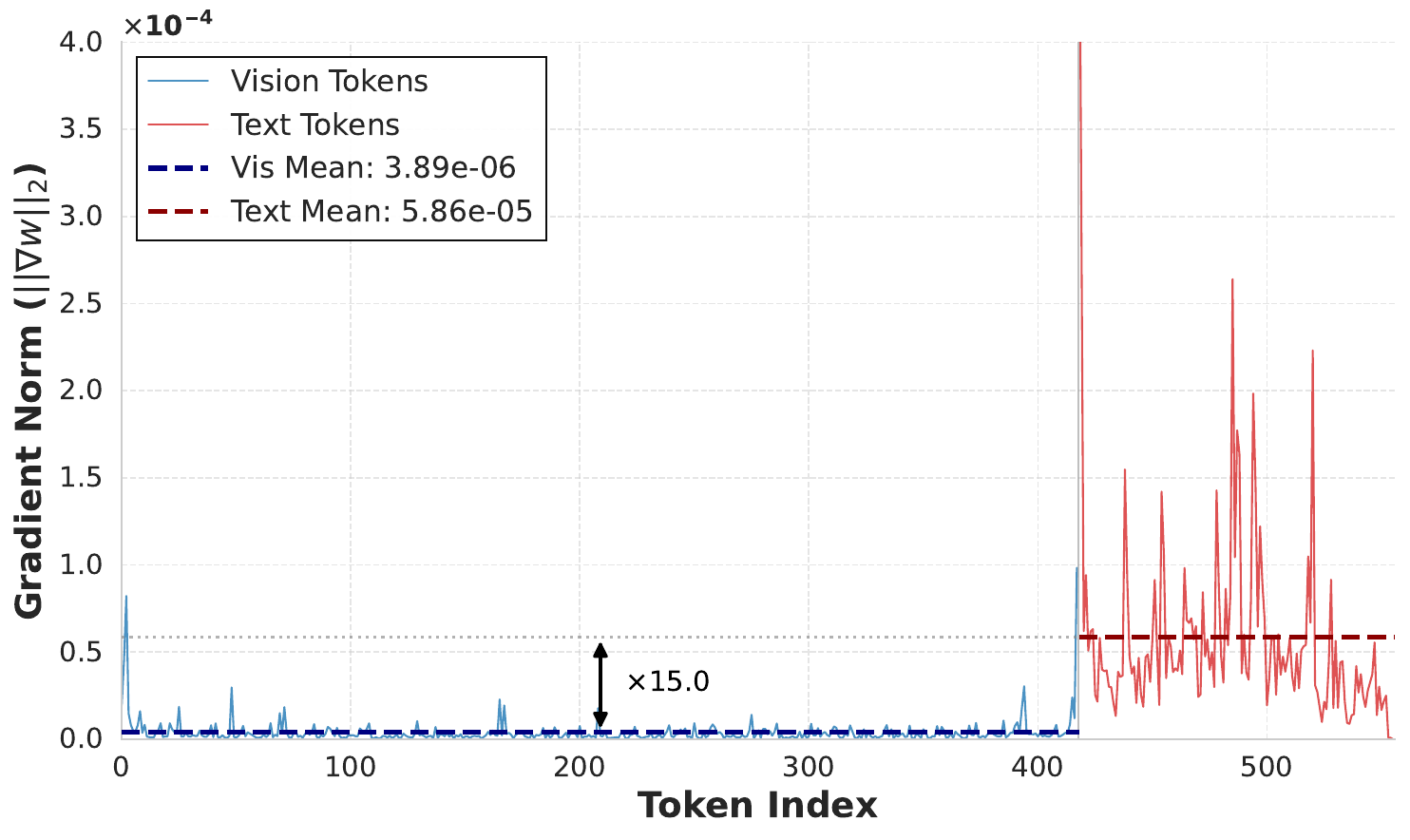}
    \vspace{-5mm}
    \caption{Detailed gradient analysis of a representative sample (Sample 88). The visualization highlights that the text-to-vision gradient ratio reaches approximately 15, confirming the dominance of textual information in the inference process.}
    \label{fig:token_grads}
    \vspace{-5mm}
\end{figure}

\vspace{-1mm}
Vision tokens, characterized by spatial redundancy, generally demonstrate lower gradient magnitudes, implying a lower sensitivity regarding their impact on inference results. In contrast, information-dense text tokens dominate the output distribution. As illustrated in Figure~\ref{fig:gradient_ratio}, Our analysis of 128 samples from the COCO \cite{lin2014coco} dataset reveals that the average gradient magnitude of text tokens exceeds that of visual tokens by a factor of \textbf{22.4}. This disparity is particularly pronounced in samples with extreme modality imbalance (e.g., containing only 5 text tokens against over 200 vision tokens), where the gradient ratio spikes drastically. This indicates that despite their scarcity, text tokens exert a disproportionately significant influence on the final generation results. Further inspection of a specific case study (Sample 88, detailed in Figure~\ref{fig:token_grads}) corroborates this observation, showing an average text-to-vision gradient ratio of \textbf{15}. Such a gradient gap across diverse scenarios highlights a fundamental sensitivity imbalance between modalities.

This discrepancy implies that employing a uniform quantization strategy across all experts ignores modality-specific sensitivities. Treating the high-impact text tokens with the same granularity as redundant vision tokens potentially compromises performance, especially in cross-modal tasks where precise language generation is paramount.

\vspace{-1mm}
\subsubsection{Experts' Heterogeneity} \label{modality-specific expert affinity analysis}
\vspace{-1mm}
To understand the internal mechanism of cross-modal processing, we investigate the routing behaviors within the MoE layers. Our analysis focuses on the activation patterns and expert preferences across different modalities, revealing three key observations regarding expert affinity.

\noindent\textbf{Intrinsic Sparsity and Load Imbalance.} 
First, we analyze the inherent sparsity of expert activation. Regardless of the calibration dataset's composition, a persistent load imbalance is observed across the expert population. Notably, even when the calibration batch size is increased to 64 (approximately 30,000 tokens), certain experts receive zero input tokens. This phenomenon indicates that the non-uniform distribution of expert utilization is an intrinsic property of the pre-trained weights rather than an artifact of data sampling. Consequently, given that a select subset of experts dominates the model's output, applying a uniform metric to measure quantization error across all experts is suboptimal.

\noindent\textbf{Heterogeneity in Token Distribution.} 
Second, the analysis of routing paths reveals a nuanced functional division of labor, characterized by the coexistence of generalist and modality-specific experts. While a subset of experts acts as universal processors activated by both visual and textual inputs, others exhibit distinct routing preferences. 

\begin{figure*}[t]
    \centering
    \begin{minipage}[b]{0.45\textwidth} 
        \centering
        \begin{minipage}[b]{0.49\linewidth}
            \centering
            \includegraphics[width=\linewidth]{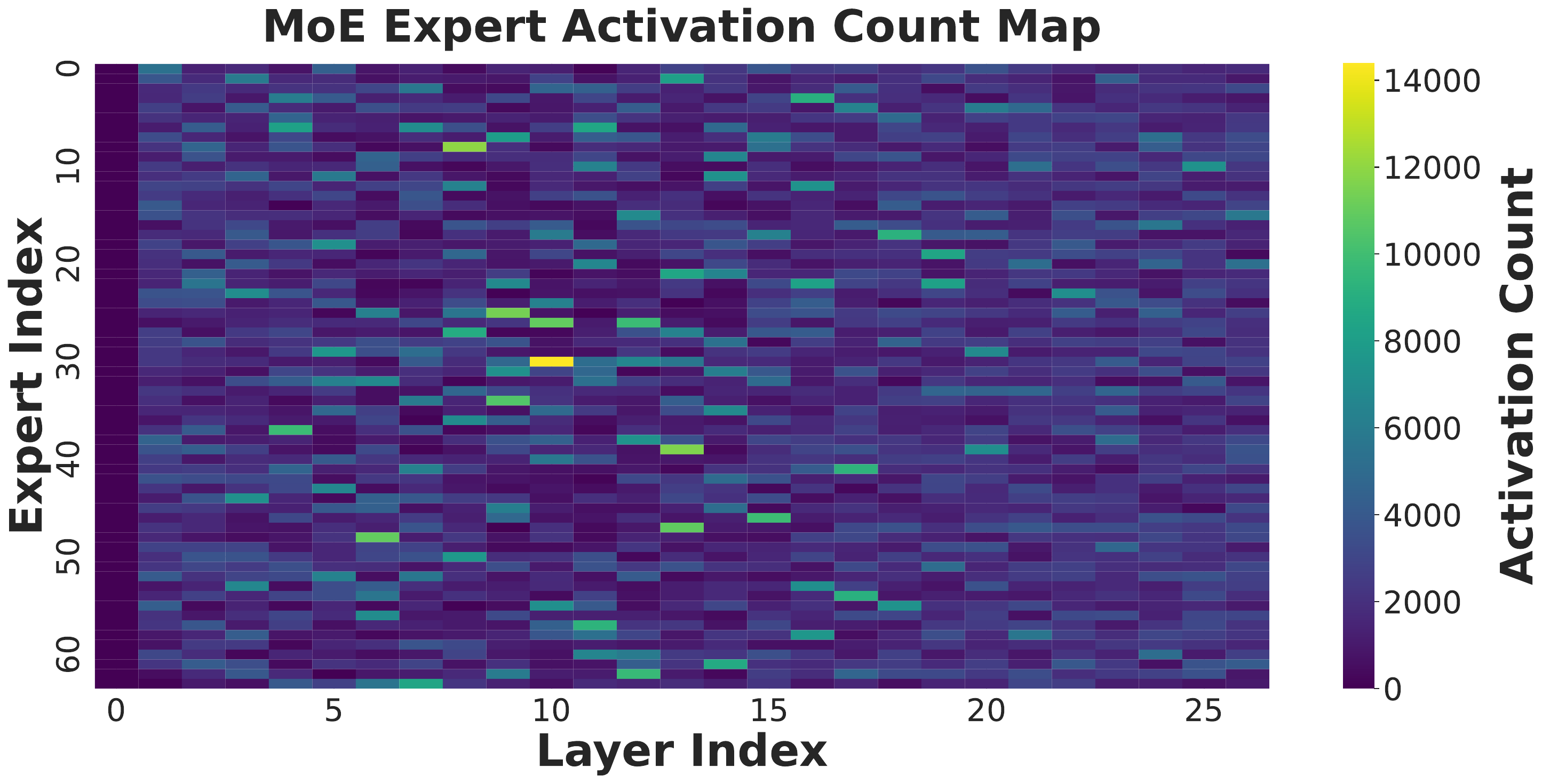}
            \vspace{-6mm}
            \subcaption{Vision Act (Set A)}
            \label{fig:vision_act_a}
        \end{minipage}
        \hfill
        \begin{minipage}[b]{0.49\linewidth}
            \centering
            \includegraphics[width=\linewidth]{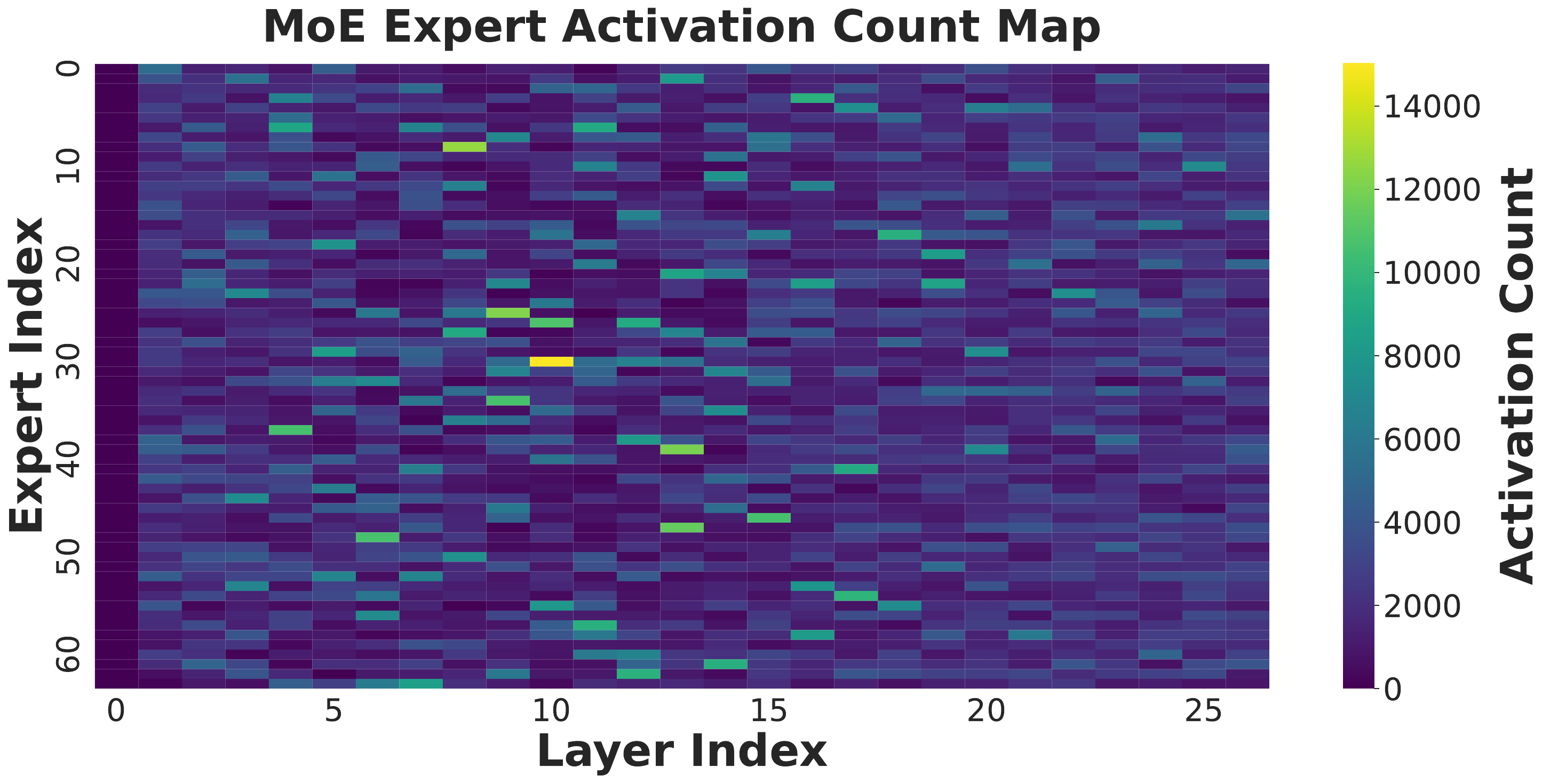}
            \vspace{-6mm}
            \subcaption{Vision Act (Set B)}
            \label{fig:vision_act_b}
        \end{minipage}

        \begin{minipage}[b]{0.49\linewidth}
            \centering
            \includegraphics[width=\linewidth]{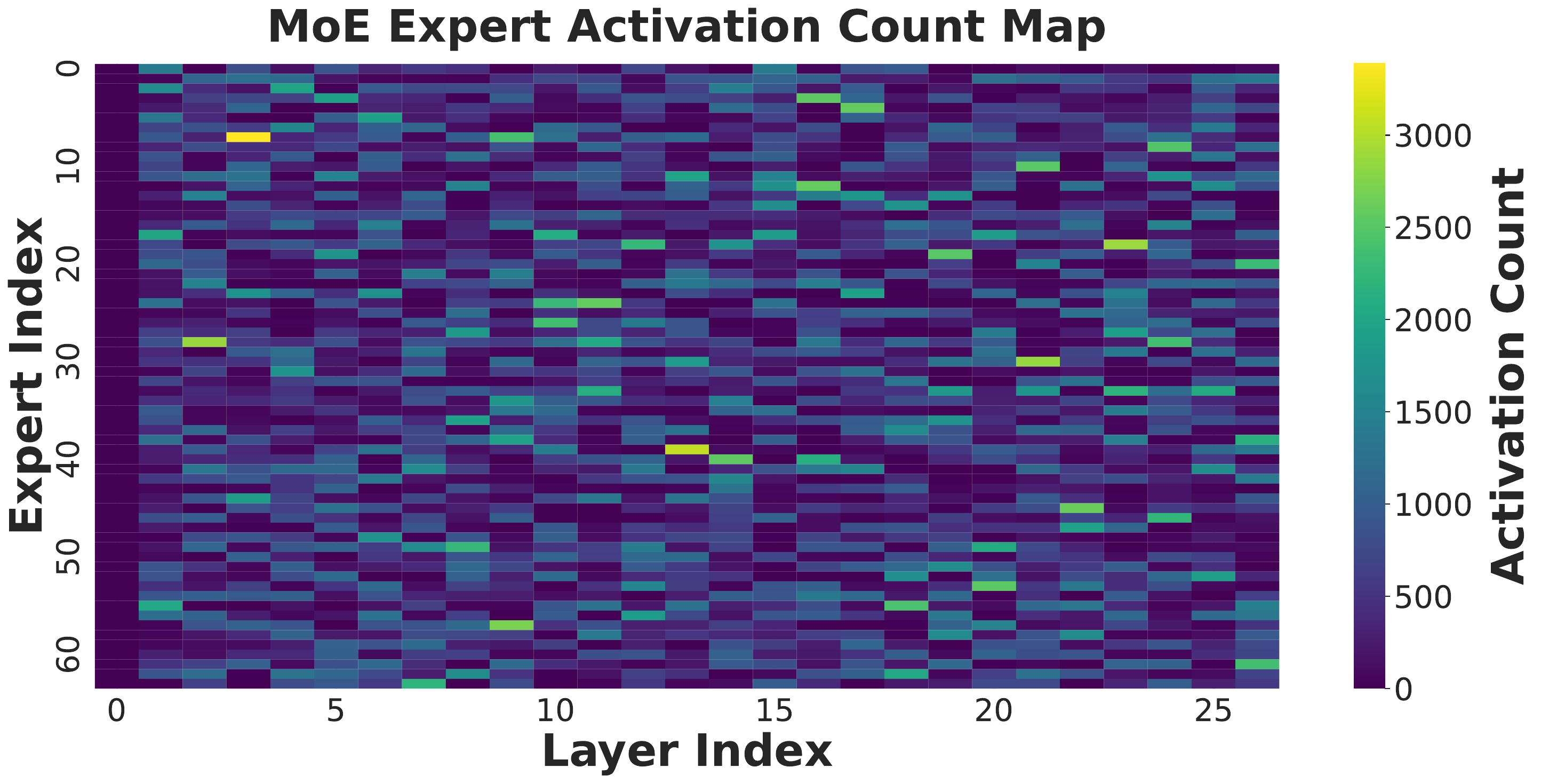}
            \vspace{-6mm}
            \subcaption{Text Act (Set A)}
            \label{fig:text_act_a}
        \end{minipage}
        \hfill
        \begin{minipage}[b]{0.49\linewidth}
            \centering
            \includegraphics[width=\linewidth]{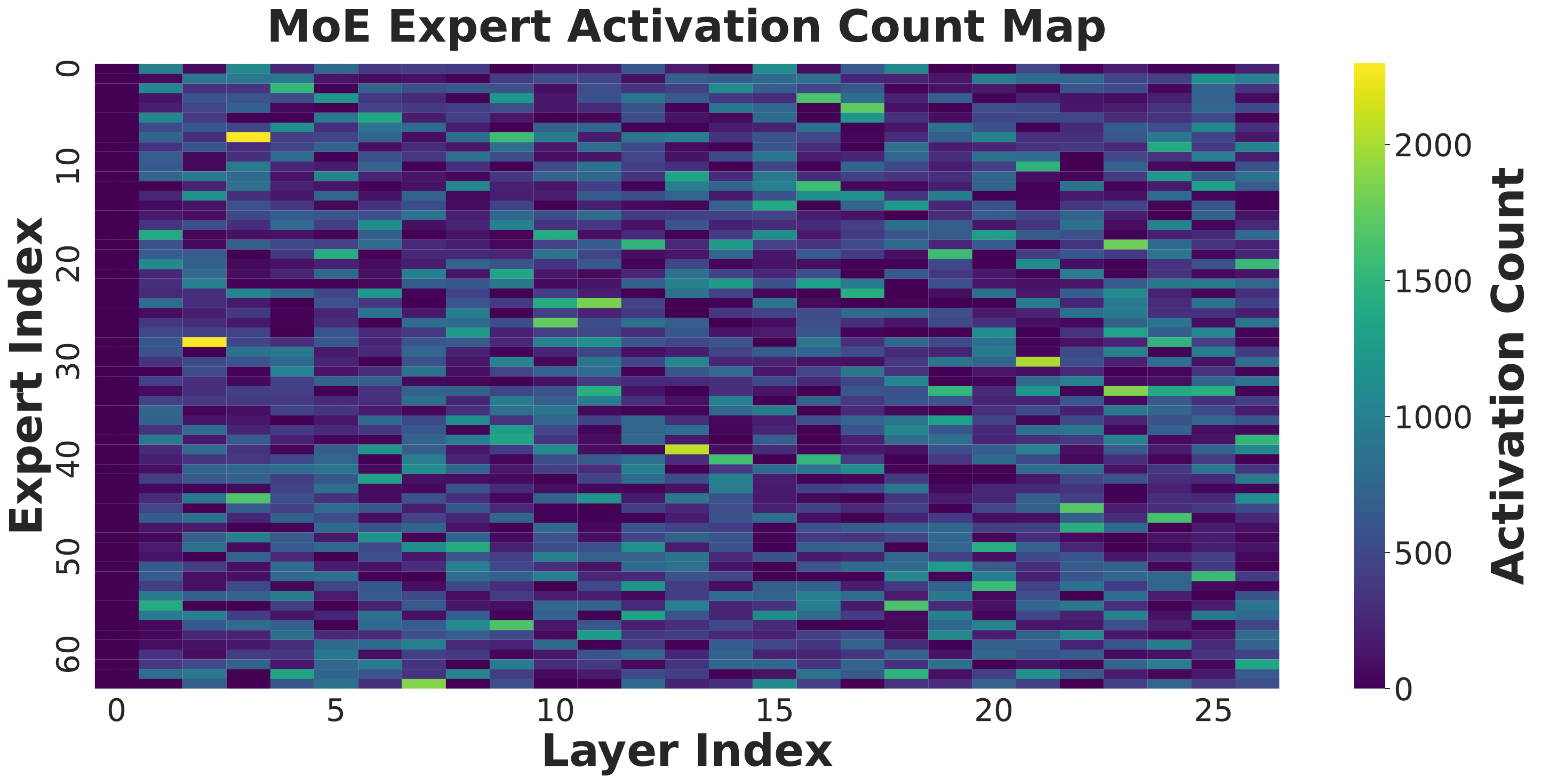}
            \vspace{-6mm}
            \subcaption{Text Act (Set B)}
            \label{fig:text_act_b}
        \end{minipage}
    \end{minipage}
    \hfill
    \begin{minipage}[b]{0.53\textwidth} 
        \centering
        \begin{subfigure}[b]{\linewidth} 
            \centering
            \includegraphics[width=\linewidth]{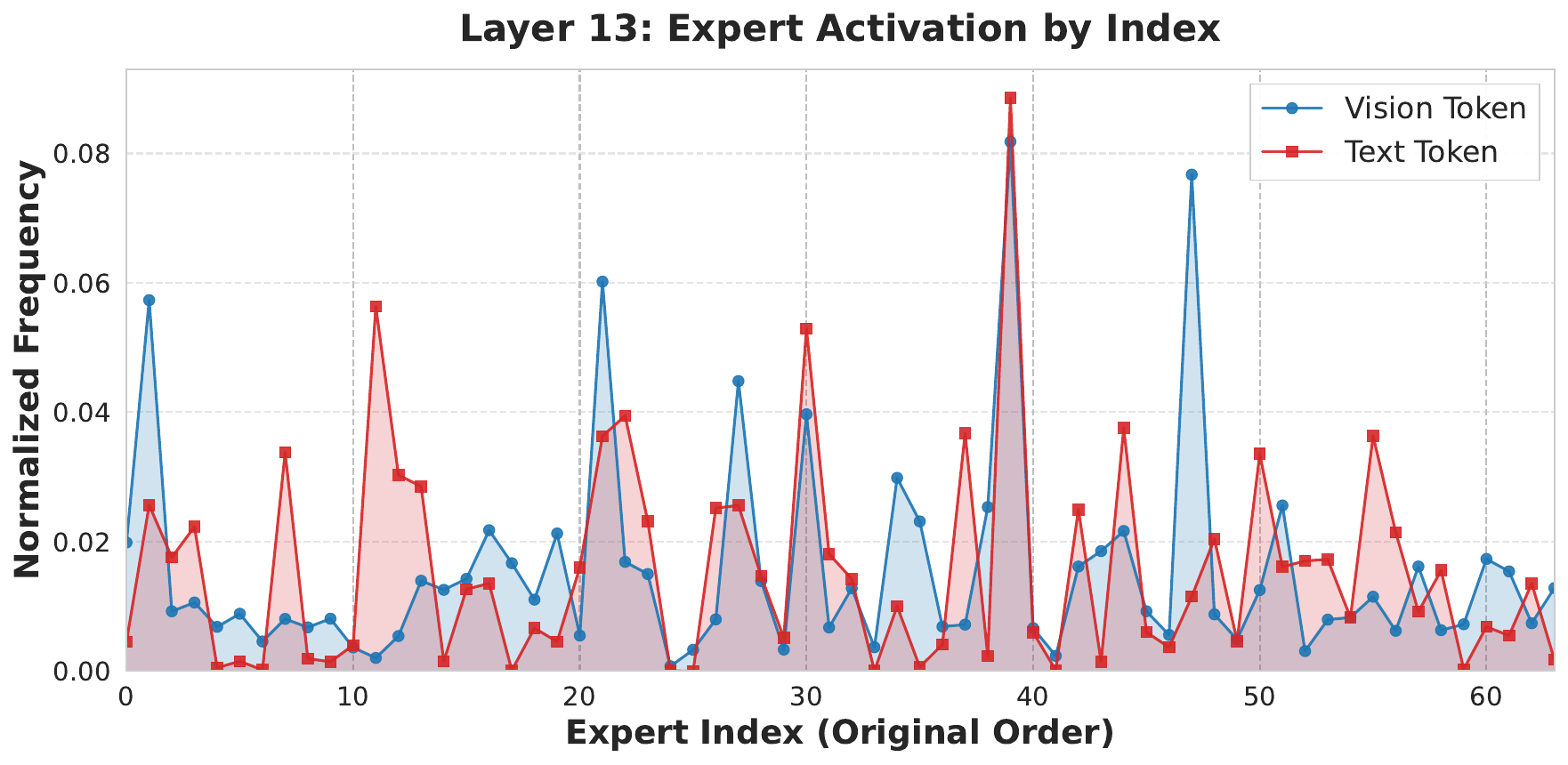}
            \vspace{-6mm}
            \caption{Experts' Activation Distribution of Layers 13 under Set A}
            \label{fig:layer_comparison}
        \end{subfigure}
    \end{minipage}

    \caption{Visualization of expert affinity patterns. Subplots (a)-(d) illustrate the distinct activation distributions for vision and text tokens across different input ranges of the \textbf{COCO dataset} \cite{lin2014coco}, highlighting how expert activation varies with different input samples while maintaining sparsity and modality-specific clustering. Subplot (e) visualizes the activation characteristics of the 13th layer in Kimi-VL-Instruct \cite{team2025kimi} under Set A, highlighting the intrinsic load imbalance.}
    \label{fig:modality_expert_analysis}
    \vspace{-5mm}
\end{figure*}

Specifically, as illustrated in Figure \ref{fig:modality_expert_analysis}(e), we observe the spontaneous emergence of specialist clusters: some are dedicated to handling spatially redundant visual features, whereas others focus exclusively on the semantic density of textual information. This distribution confirms that the model allocates capacity both for cross-modal alignment and for modality-specific processing. The distinct activation frequencies and magnitudes across these clusters underscore the uneven contribution of different experts. 
Consequently, it is imperative to assign quantization error weights according to these modality-dependent activation characteristics, ensuring that the optimization strategy adapts to the specific functional role of each expert. 

\noindent\textbf{Routing Bias and Decisive Experts.} 
Finally, the router outputs demonstrate significant bias, where a small fraction of experts plays a decisive role in the model's output. The routing probability distribution is highly skewed; for any given token, the router assigns high confidence scores to only a few experts (the top-$k$ selection), while the affinity for the remaining experts approaches zero. As indicated by the gradient magnitude analysis, these high-affinity experts dominate the inference outcome, whereas the influence of the non-selected experts is negligible. This implies that preserving the precision of these decisive experts is critical for maintaining model performance.

\vspace{-2mm}
\subsection{Modality-Expert-Aware Quantization}\label{Modality-Expert-Aware Quantization}
\vspace{-1mm}
Building upon the analysis in \cref{modality experts heterogeneity analysis}, we have established that experts within MoE layers exhibit significant heterogeneity in both activation frequency and modality-specific sensitivity.Representative PTQ frameworks, such as AWQ \cite{lin_awq_2024}, typically determine optimal quantization parameters via a grid search aimed at minimizing generic reconstruction loss. However, such a search process treats all experts indiscriminately, failing to account for their varying importance during inference.

To address this, we propose a Modality-Expert-Aware Quantization method that incorporates expert importance into the error minimization objective.  

\textbf{Quantifying Expert Importance.} First, we define an importance score $S_i$ for the $i$-th expert as a balanced measure of its contribution across modalities. Due to the inherent modality imbalance in MoE VLMs, where vision tokens significantly outnumber text tokens, a raw frequency count would disproportionately favor vision-dominant experts.

Therefore, we introduce a comprehensive normalization mechanism to better align the magnitude of contributions from different modalities. The importance score $S_i$ is formulated as a weighted summation: 
\begin{equation}
    S_i = \gamma \cdot N_{i}^{\text{text}} + \beta \cdot N_{i}^{\text{vis}},
    \label{eq:expert_score}
\end{equation}
\noindent where $N_{i}^{\text{text}}$ and $N_{i}^{\text{vis}}$ denote the number of text and visual tokens routed to the $i$-th expert, respectively. Let $T_{\text{text}}$ and $T_{\text{vis}}$ represent the total number of text and vision tokens in the calibration set, respectively. The coefficient $\beta = T_{\text{text}} / T_{\text{vis}}$ serves as a \textbf{quantity normalization factor}, scaling down the frequent visual activations to be comparable with textual counts. The coefficient $\gamma = \|\nabla_{\text{text}}\| / \|\nabla_{\text{vis}}\|$ acts as a \textbf{quality sensitivity factor}, reflecting the higher gradient impact of text tokens. This formulation ensures that the resulting score is driven by the semantic significance of the tokens rather than their raw frequency, protecting the decisive experts responsible for logical reasoning.

\textbf{Importance-Aware Optimization Objective.} As mentioned above, established PTQ strategies like AWQ \cite{lin_awq_2024} optimize quantization scales through grid-based search, treating the MoE layer as a monolithic entity. These methods apply a uniform optimization target across the model. Specifically within the MoE module, this search process is equivalent to minimizing an unweighted summation of reconstruction errors across all experts:
\begin{equation}
    \mathcal{L}_{\text{Standard}} = \sum_{i=1}^{M} \| \mathbf{W}_i \mathbf{X}_i - \hat{\mathbf{W}}_i \mathbf{X}_i \|_F^2,
\end{equation}
\noindent where $\mathbf{W}_i$ and $\hat{\mathbf{W}}_i$ denote the full-precision and quantized weights of the $i$-th expert, respectively, $\mathbf{X}_i$ represents the input tokens routed to that expert, and $M$ is the total number of experts. This formulation treats rarely used experts and critical experts indistinguishably, failing to account for their varying contributions to the model's final inference results.

In contrast, our proposed method redefines the optimization objective by introducing the importance score $S_i$ (defined in Eq.~\ref{eq:expert_score}) as a weighting factor. The weighted quantization error is expressed as the following equation:
\begin{equation}
    \mathcal{L}_{\text{Weighted}} = \sum_{i=1}^{M} S_i \cdot \| \mathbf{W}_i \mathbf{X}_i - \hat{\mathbf{W}}_i \mathbf{X}_i \|_F^2.
    \label{eq:weighted_error}
\end{equation}
By minimizing $\mathcal{L}_{\text{Weighted}}$ during the search for optimal quantization parameters, quantization noise is preferentially suppressed in experts that exhibit high activation frequencies and high modality sensitivities. This ensures that the quantization parameters are calibrated to preserve the functionality of the most influential experts, thereby accurately reflecting the true impact of quantization on the model's overall performance on downstream tasks.
\vspace{-2mm}
\subsection{Modality-Affinity-Aware Quantization}\label{Modality-Affinity-Aware Quantization}
\vspace{-1mm}
Some other representative post-training quantization frameworks, such as GPTQ \cite{frantar_gptq_2023}, typically rely on second-order information to determine the optimal quantization parameters. Specifically, they utilize the Hessian matrix $H$, approximations of which are computed via input calibration data $X$ as $H = 2XX^\top$. This formulation implicitly assumes that all input tokens contribute equally to the reconstruction error, treating the optimization landscape as uniform across the sequence dimension.

\textbf{Affinity and Modality Imbalance.} However, as analyzed in \cref{modality experts heterogeneity analysis}, this assumption of uniformity is not suitable for MoE VLMs. The input tokens exhibit significant variance in their interaction with experts: 
(1) Routing Diversity: Tokens possess varying levels of affinity with specific experts, determined by the router's output probabilities; 
(2) Modality Sensitivity: Text tokens, despite being fewer in number, carry higher gradient densities and information value compared to spatially redundant vision tokens. 

Directly applying a uniform Hessian calculation to the MoE layers fails to account for these nuances, potentially causing the neglection of critical semantic information.

\textbf{Enhanced Hessian Matrix.} To address this, we propose an effective Modality-Affinity-Aware formulation for the Hessian matrix. We introduce a token-wise importance weighting mechanism. Let $X \in \mathbb{R}^{d \times N}$ denote the input tokens specifically routed to the current expert, where $N$ is the number of such tokens. We reconstruct the Hessian matrix $\tilde{H}$ by scaling the contribution of each token according to its modality-specific affinity: 
\begin{equation}
    \tilde{H} = (X \cdot \sqrt{\mathbf{C}})(X \cdot \sqrt{\mathbf{C}})^\top = X \mathbf{C} X^\top,
    \label{eq:weighted_hessian}
\end{equation}
\noindent where $\mathbf{C} \in \mathbb{R}^{N \times N}$ is a diagonal matrix representing the importance weight for each token. For the $j$-th token $x_j$, the corresponding diagonal element $c_j$ is defined as:
\begin{equation}
    c_j = p_{j} \cdot \alpha_j, \quad \text{where } \alpha_j = 
    \begin{cases} 
    \gamma & x_j \text{ is text token}, \\
    1 & x_j \text{ is vision token}.
    \end{cases}
\end{equation}
\noindent Here, $p_{j}$ denotes the affinity between token $j$ and the target expert. The term $\gamma$ represents the gradient scaling factor, formally defined as the ratio of gradient magnitudes between textual and visual modalities: $\gamma = \|\nabla_{\text{text}}\| / \|\nabla_{\text{vis}}\|$.

By incorporating $\mathbf{C}$ into the Hessian computation, the quantization objective is dynamically re-weighted. Tokens with high router affinity and high modality sensitivity exert a larger influence on the structure of $\tilde{H}$. Consequently, when the Inverse Hessian is applied to update the weights, the algorithm prioritizes preserving the accuracy of these high-impact tokens, thereby maximizing the model's performance retention under low-bit settings.

\begin{table*}[!t]
\centering
\caption{Main comparison results of Kimi-VL-Instruct and Qwen3-VL-30B-A3B-Instruct under 3-bit (W3) and 4-bit (W4) weight only quantization. We report the zero-shot accuracy (\%) for all tasks. The best results for each bit-width are highlighted in \textbf{bold}. The improvement of VEQ-MA over the best baseline method is marked in parentheses. Abbreviations: InfoV: InfoVQA, TextV: TextVQA, RWQA: RealWorldQA, SciQA: ScienceQA, Viz: VizWiz, MMB: MMBench, MME-R: MME-RealWorld.}
\label{tab:main_results}

\definecolor{graybg}{gray}{0.95}
\renewcommand{\arraystretch}{1.2}

\resizebox{\textwidth}{!}{%
\begin{tabular}{lllcccccccccc}
\toprule
\rowcolor{colorhead}
  &  &  & \multicolumn{9}{c}{Benchmarks} &  \\
\rowcolor{colorhead}
 \multirow{-2}{*}{Model} & \multirow{-2}{*}{Bit} & \multirow{-2}{*}{Method} & MMMU & AI2D & InfoV & TextV & RWQA & SciQA & Viz & MMB & MME-R & \multirow{-2}{*}{Avg.} \\
\midrule

\multirow{14}{*}{\shortstack[l]{Kimi-VL-\\Instruct}} 
 & \textbf{BF16} & - & 51.11 & 83.65 & 83.38 & 85.93 & 66.41 & 92.10 & 69.00 & 82.73 & 58.22 & 74.73 \\
\cmidrule(lr){2-13} \addlinespace[3pt]

 & \multirow{6}{*}{W3} 
 & RTN & 37.00 & 69.14 & 59.68 & 74.58 & 56.47 & 79.06 & 58.51 & 66.06 & 46.13 & 60.74 \\
 & & AWQ & 41.78 & 70.05 & 61.49 & 74.04 & 58.17 & 81.84 & 61.37 & 71.13 & 50.44 & 63.37 \\
 & & MBQ & 39.76 & 70.89 & 58.86 & 74.35 & 58.74 & 81.82 & 60.87 & 71.13 & 49.91 & 62.93 \\
 & & GPTQ & 40.33 & 75.49 & 64.14 & 64.30 & 58.82 & 83.87 & 55.61 & 73.02 & 45.41 & 62.33 \\
 & & \cellcolor{graybg}VEQ-ME (ours) & \cellcolor{graybg}\textbf{44.56} & \cellcolor{graybg}71.57 & \cellcolor{graybg}62.33 & \cellcolor{graybg}73.36 & \cellcolor{graybg}\textbf{58.95} & \cellcolor{graybg}82.93 & \cellcolor{graybg}58.90 & \cellcolor{graybg}71.91 & \cellcolor{graybg}\textbf{51.46} & \cellcolor{graybg}64.00 \\
 & & \cellcolor{graybg}VEQ-MA (ours) & \cellcolor{graybg}42.56 & \cellcolor{graybg}\textbf{75.65} & \cellcolor{graybg}\textbf{64.48} & \cellcolor{graybg}\textbf{78.30} & \cellcolor{graybg}58.30 & \cellcolor{graybg}\textbf{84.85} & \cellcolor{graybg}\textbf{63.10} & \cellcolor{graybg}\textbf{73.42} & \cellcolor{graybg}48.03 & \cellcolor{graybg}\textbf{65.41} \color[HTML]{228B22}{\scriptsize{(+2.04)}} \\
\cmidrule(lr){2-13} \addlinespace[3pt]

 & \multirow{6}{*}{W4} 
 & RTN & 48.44 & 82.38 & 80.16 & 84.60 & \textbf{66.80} & 90.83 & \textbf{69.17} & 80.67 & 52.48 & 72.84 \\
 & & AWQ & 49.00 & 81.74 & 79.01 & 83.37 & 66.27 & 90.92 & 69.07 & 80.58 & 53.70 & 72.63 \\
 & & MBQ & 48.89 & 81.70 & 78.49 & 83.18 & 64.58 & 90.29 & 69.03 & 80.93 & 55.08 & 72.46 \\
 & & GPTQ & 50.00 & 82.32 & 80.04 & 84.63 & 64.84 & 91.39 & 68.86 & 81.44 & 53.14 & 72.96 \\
 & & \cellcolor{graybg}VEQ-ME (ours) & \cellcolor{graybg}49.22 & \cellcolor{graybg}81.41 & \cellcolor{graybg}79.75 & \cellcolor{graybg}83.69 & \cellcolor{graybg}66.36 & \cellcolor{graybg}91.06 & \cellcolor{graybg}69.17 & \cellcolor{graybg}80.84 & \cellcolor{graybg}54.20 & \cellcolor{graybg}72.86 \\
 & & \cellcolor{graybg}VEQ-MA (ours) & \cellcolor{graybg}\textbf{50.30} & \cellcolor{graybg}\textbf{82.55} & \cellcolor{graybg}\textbf{80.37} & \cellcolor{graybg}\textbf{84.71} & \cellcolor{graybg}66.41 & \cellcolor{graybg}\textbf{91.56} & \cellcolor{graybg}68.91 & \cellcolor{graybg}\textbf{81.67} & \cellcolor{graybg}\textbf{55.61} & \cellcolor{graybg}\textbf{73.57} \color[HTML]{228B22}{\scriptsize{(+0.61)}} \\

\midrule

\multirow{14}{*}{\shortstack[l]{Qwen3-VL-\\30B-A3B-\\Instruct}} 
 & \textbf{BF16} & - & 73.67 & 86.27 & 81.43 & 81.31 & 65.49 & 93.52 & 83.27 & 85.91 & 59.92 & 78.98 \\
\cmidrule(lr){2-13} \addlinespace[3pt]

 & \multirow{6}{*}{W3} 
 & RTN & 57.33 & 77.10 & 44.96 & 68.64 & 43.01 & 89.08 & 64.07 & 78.60 & \textbf{45.73} & 63.17 \\
 & & AWQ & 58.89 & 73.61 & 44.62 & 67.93 & 45.36 & 88.40 & 61.75 & 77.06 & 42.38 & 62.22 \\
 & & MBQ & 50.56 & 71.44 & 40.15 & 64.23 & 53.82 & 87.31 & 59.48 & 76.46 & 45.27 & 60.97 \\

 & & GPTQ & 64.89 & 74.11 & \textbf{47.27} & 71.82 & 54.25 & 88.23 & 63.39 & 69.39 & 44.08 & 64.16 \\
 & & \cellcolor{graybg}VEQ-ME (ours) & \cellcolor{graybg}60.56 & \cellcolor{graybg}73.38 & \cellcolor{graybg}44.95 & \cellcolor{graybg}69.91 & \cellcolor{graybg}53.73 & \cellcolor{graybg}88.40 & \cellcolor{graybg}62.47 & \cellcolor{graybg}77.75 & \cellcolor{graybg}44.24 & \cellcolor{graybg}63.93 \\
 & & \cellcolor{graybg}VEQ-MA (ours) & \cellcolor{graybg}\textbf{65.89} & \cellcolor{graybg}\textbf{79.15} & \cellcolor{graybg}47.14 & \cellcolor{graybg}\textbf{72.74} & \cellcolor{graybg}\textbf{54.77} & \cellcolor{graybg}\textbf{89.55} & \cellcolor{graybg}\textbf{69.46} & \cellcolor{graybg}\textbf{82.30} & \cellcolor{graybg}43.25 & \cellcolor{graybg}\textbf{67.14} \color[HTML]{228B22}{\scriptsize{(+3.09)}} \\
\cmidrule(lr){2-13} \addlinespace[3pt]

 & \multirow{6}{*}{W4} 
 & RTN & 63.11 & \textbf{83.33} & 62.62 & 79.77 & 45.23 & 91.51 & 71.08 & 85.13 & 57.31 & 70.89 \\
 & & AWQ & 65.20 & 81.22 & 58.32 & 77.40 & 63.14 & 91.06 & 69.73 & 85.22 & 57.31 & 72.07 \\
 & & MBQ & 69.67 & 81.41 & 59.36 & 78.99 & 64.31 & 91.51 & 70.78 & 83.68 & 56.10 & 72.87 \\

 & & GPTQ & \textbf{72.67} & 82.93 & 62.54 & 80.18 & 53.46 & \textbf{93.23} & 71.32 & 84.53 & 57.96 & 73.20 \\
 & & \cellcolor{graybg}VEQ-ME (ours) & \cellcolor{graybg}68.78 & \cellcolor{graybg}82.47 & \cellcolor{graybg}59.53 & \cellcolor{graybg}78.57 & \cellcolor{graybg}\textbf{65.32} & \cellcolor{graybg}91.61 & \cellcolor{graybg}70.97 & \cellcolor{graybg}83.56 & \cellcolor{graybg}56.79 & \cellcolor{graybg}73.07 \\
 & & \cellcolor{graybg}VEQ-MA (ours) & \cellcolor{graybg}71.56 & \cellcolor{graybg}82.95 & \cellcolor{graybg}\textbf{62.86} & \cellcolor{graybg}\textbf{81.10} & \cellcolor{graybg}61.70 & \cellcolor{graybg}92.64 & \cellcolor{graybg}\textbf{72.42} & \cellcolor{graybg}\textbf{85.88} & \cellcolor{graybg}\textbf{58.12} & \cellcolor{graybg}\textbf{74.36} \color[HTML]{228B22}{\scriptsize{(+1.16)}} \\

\bottomrule
\end{tabular}
}
\vspace{-3mm}
\end{table*}

\vspace{-2mm}
\section{Experiments}
\vspace{-1mm}
\subsection{Experiment setting}
\vspace{-1mm}
We conduct comprehensive experiments to evaluate the efficacy of our proposed quantization method.

\vspace{-1mm}
\noindent\textbf{Model and Benchmarks.}
We choose to utilize Kimi-VL-Instruct \cite{team2025kimi} and Qwen3-VL-30B-A3B-Instruct \cite{Qwen3-VL} for experiments. To ensure a robust assessment of multimodal capabilities, we employ a diverse set of widely recognized benchmarks, including MMMU \cite{yue_mmmu_2024} and MMBench \cite{liu2024mmbench} for multi-discipline reasoning, AI2D \cite{kembhavi_diagram_2016} for diagram understanding, InfoVQA \cite{mathew_infographicvqa_2021} and TextVQA \cite{singh2019towards} for OCR-related visual question answering, as well as MME-RealWorld \cite{zhang2024mme}, RealWorldQA \cite{realworldqa}, ScienceQA \cite{lu2022learn} and VizWiz-VQA \cite{gurari2019vizwiz} to cover real-world and scientific scenarios.

\vspace{-1mm}
\noindent\textbf{Baselines.} 
We compare our method against the full-precision version and several established quantization baselines to demonstrate its superiority:
1) BF16: The original model in Bfloat16 precision, serving as the performance upper bound;
2) RTN (Round-to-Nearest): A naive baseline that quantizes weights by rounding them to the nearest grid point;
3) Advanced PTQ Frameworks: We include GPTQ \cite{frantar_gptq_2023} and AWQ \cite{lin_awq_2024}, which are widely adopted as standard baselines for LLM compression, alongside MBQ \cite{li2025mbq}, the current SOTA method specifically tailored for VLM quantization.

\vspace{-1mm}
\noindent\textbf{Implementation Details.} 
Our evaluation pipeline is built upon the open-source lmms-eval \cite{lmms_eval2024} framework, a standardized toolkit designed for the rigorous evaluation of Large Multimodal Models. For the inference backend, we employ SGLang \cite{zheng2024sglang}, a high-throughput serving engine optimized for MoE architectures to make inference efficient. During evaluation process, model outputs are generated and then compared against the standard ground-truth answers for each benchmark to compute the final accuracy scores. 

\vspace{-2mm}
\subsection{Main Results}
\vspace{-1.5mm}
We conduct a comprehensive evaluation of our proposed method against state-of-the-art baselines on the Kimi-VL-Instruct \cite{team2025kimi} and Qwen3-VL-30B-A3B-Instruct \cite{Qwen3-VL} models. For clarity in the following analysis, we denote the implementation of \textbf{Modality-Expert-Aware Quantization} based on AWQ \cite{lin_awq_2024} as VEQ-ME, while the version incorporating \textbf{Modality-Affinity-Aware Quantization} based on GPTQ is referred to as VEQ-MA. To fully assess the robustness of quantization, we perform experiments under both 4-bit (W4) and 3-bit (W3) weight quantization settings. The detailed comparison results across seven multimodal benchmarks are presented in Table~\ref{tab:main_results}.

\vspace{-1mm}
\noindent\textbf{Performance under W4 Setting.} As shown in the upper section of Table~\ref{tab:main_results}, most methods maintain robust performance under the 4-bit quantization setting. Specifically, for the Kimi-VL-Instruct model, established baselines such as AWQ \cite{lin_awq_2024}, MBQ \cite{li2025mbq} and GPTQ \cite{frantar_gptq_2023}, along with our proposed VEQ, recover nearly 98\% of the average BF16 accuracy. The relatively marginal performance gap among different quantization strategies suggests that 4-bit precision offers sufficient capacity to represent MoE weights without incurring catastrophic information loss.

\noindent\textbf{Performance under W3 Setting.}
The distinction between methods becomes significantly more pronounced in the aggressive 3-bit setting. As shown in the lower section of Table~\ref{tab:main_results}, traditional baselines such as RTN and AWQ \cite{lin_awq_2024} suffer from severe degradation, particularly on reasoning-intensive tasks like MMMU \cite{yue_mmmu_2024} and fine-grained visual tasks like InfoVQA \cite{mathew_infographicvqa_2021}. This suggests that the uniform quantization assumptions fail when the bit-width is extremely limited. It also proves that ignoring the heterogeneity can lead to errors under extreme compression. In contrast, VEQ demonstrates exceptional robustness. By explicitly modeling expert importance and modality affinity, VEQ significantly outperforms the baselines. For instance, on TextVQA \cite{singh2019towards}, VEQ achieves a gain of \textbf{21.4\%} compared to the original quantization method on Kimi-VL-Instruct. These results validate that protecting the decisive experts and differentiating modality sensitivities are critical in low-bit regimes.

\subsection{Ablation Studies}
\vspace{-1mm}
To verify the effectiveness and robustness of our proposed method, we conduct a two-fold ablation study. First, we evaluate the contribution of each component to the overall performance on downstream tasks. Second, we perform a hyperparameter sensitivity analysis on a randomly extracted validation set to justify our parameter selection. In addition, we report all ablation results under the same quantization configuration to ensure a fair comparison. We further keep the calibration data and evaluation protocol fixed across settings to isolate the effect of each component. 
\begin{figure}[!t]
    \centering
    \begin{subfigure}[b]{0.48\columnwidth}
        \centering
        \includegraphics[width=\linewidth]{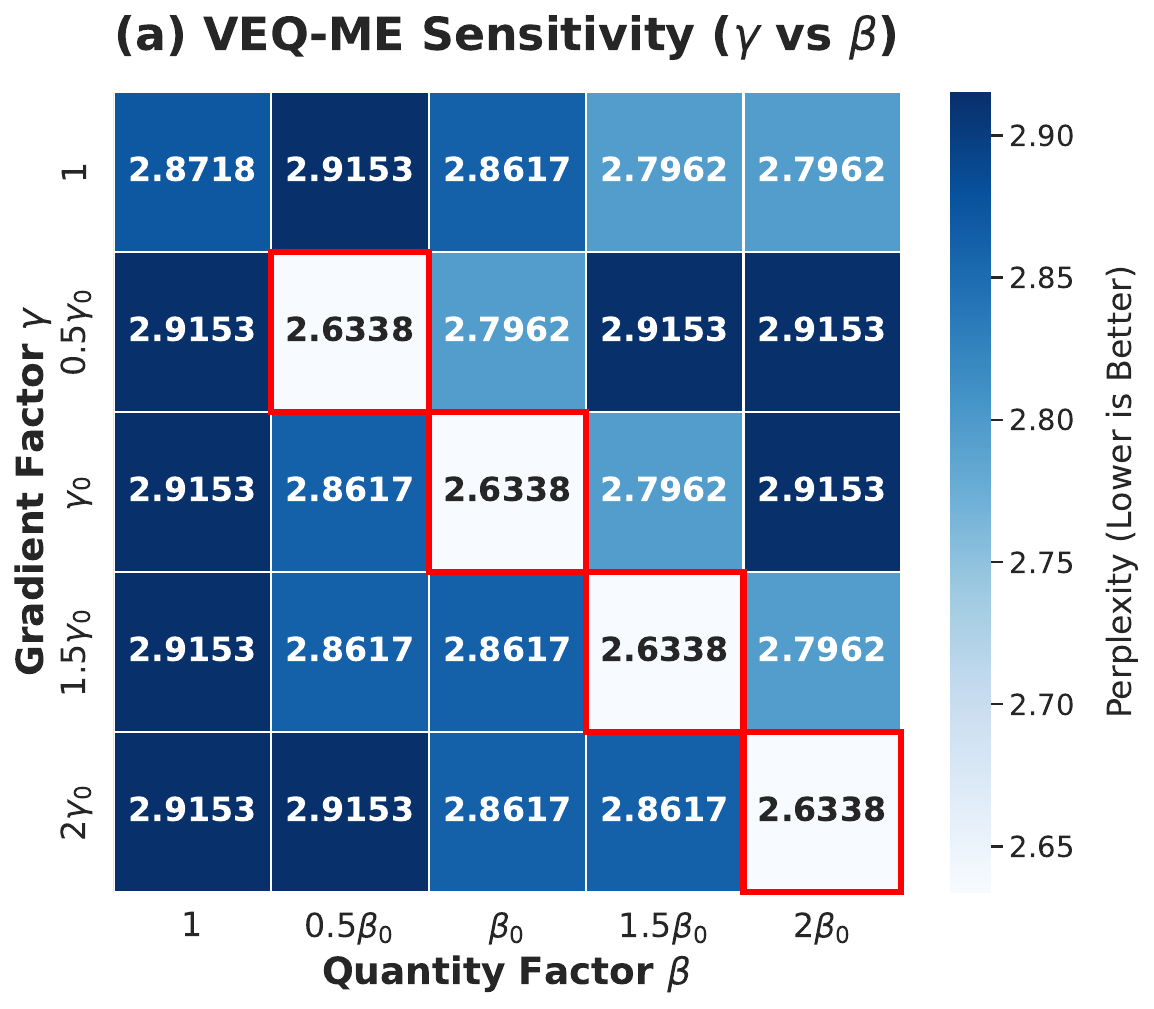}
        \caption{VEQ-ME Sensitivity.}
        \label{fig:veq_plus_landscape}
    \end{subfigure}
    \hfill
    \begin{subfigure}[b]{0.48\columnwidth}
        \centering
        \includegraphics[width=\linewidth]{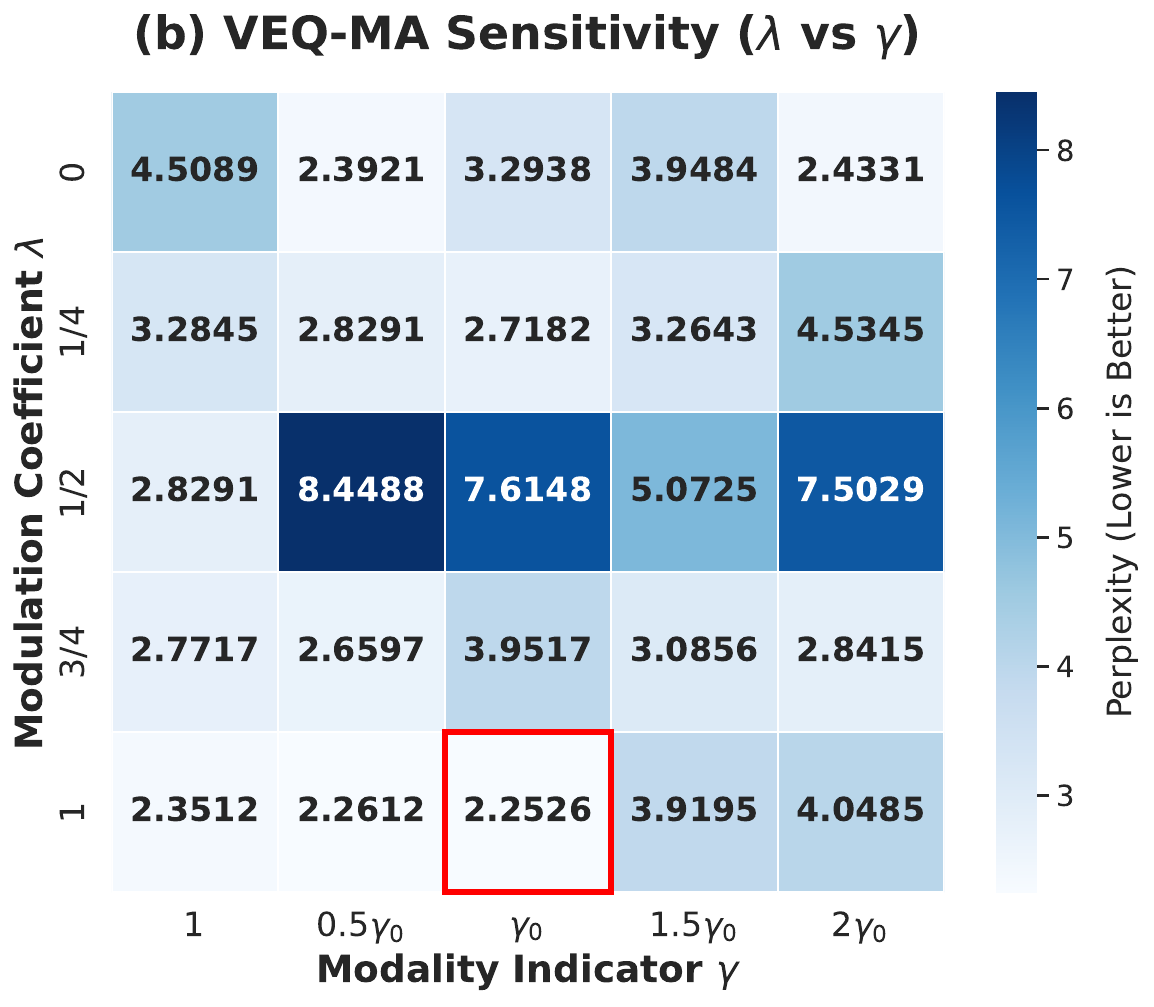}
        \caption{VEQ-MA Sensitivity.}
        \label{fig:veq_pp_landscape}
    \end{subfigure}

    \vspace{-1mm}
    \caption{Visual analysis of parameter sensitivity regarding validation PPL. (a) VEQ-ME: It confirms the scale invariance of our method, where maintaining the relative ratio ensures consistent minimization of quantization error.
    (b) VEQ-MA: The results shows reducing $\lambda$ generally results in an increase in PPL, validating that the router confidence is important for accurate quantization.}
    \label{fig:ablation_heatmap}
    \vspace{-2mm}
\end{figure}

\vspace{-1mm}
\noindent\textbf{Component Effectiveness on Downstream Tasks.}
We focus on the two core formulations: the Expert Importance Score in VEQ-ME and the Modality-Affinity-Aware Hessian in VEQ-MA. We set the hyperparameters to their default optimal values and measure the performance drop when specific components are disabled. 

\noindent\textit{1) Impact of Modality-Expert Importance (VEQ-ME).}
The importance score is formulated as $S_i = \gamma N_i^{text} + \beta N_i^{vision}$. We investigate the necessity of the gradient scaling factor $\gamma$ and the quantity normalization factor $\beta$:
\begin{itemize}
\vspace{-2mm}
    \item \textbf{w/o $\gamma$ ($\gamma=1$):} Removing the gradient scale treats visual and textual importance purely based on token quantity. As shown in Table~\ref{tab:ablation_veq1}, this leads to a noticeable drop in text-heavy reasoning tasks, confirming that text tokens require higher sensitivity weights.
    \vspace{-1mm}
    \item \textbf{w/o $\beta$ ($\beta=1$):} Ignoring the quantity gap allows the vast number of vision tokens to dominate the score. The results indicate that this variation degrades performance, as the router becomes biased toward spatially redundant visual features.
    \vspace{-1mm}
\end{itemize}

\begin{table}[t]
\centering
\caption{Ablation study of Modality-Expert Importance (VEQ-ME) on downstream tasks. $\gamma$: Gradient factor; $\beta$: Quantity factor. }
\label{tab:ablation_veq1}
\resizebox{\columnwidth}{!}{
\begin{tabular}{lcccc}
\toprule
\rowcolor{colorhead}
Config & MMMU & InfoVQA & ScienceQA & Avg. \\
\midrule
w/o $\gamma$ ($\gamma=1$) & 43.44 & 60.39 & 82.24 & 62.02 \\
w/o $\beta$ ($\beta=1$) & 43.89 & 60.34 & 82.67 & 62.30 \\
\textbf{VEQ-ME (Full)} & \textbf{44.56} & \textbf{62.33} & \textbf{82.93} & \textbf{63.27} \\
\bottomrule
\end{tabular}
}
\vspace{-4mm}
\end{table}

\vspace{-1.5mm}
\noindent\textit{2) Impact of Modality-Affinity Awareness (VEQ-MA).}
For affinity-aware quantization, the token weight is defined as $c_j = p_j \cdot \alpha_j$. We examine the roles of affinity $p_j$ and the modality indicator $\alpha_j$:
\begin{itemize}
\vspace{-2mm}
    \item \textbf{w/o $p$ ($p=1$):} Ignoring router affinity logits treats all tokens routed to an expert as equally important. Table~\ref{tab:ablation_veq2} shows that this leads to suboptimal results, proving that tokens with higher routing confidence are more representative.
    \vspace{-1mm}
    \item \textbf{w/o $\alpha$ ($\alpha=1$):} Removing modality-specific re-weighting during Hessian calibration causes a performance decline, confirming that distinguishing between information-dense text tokens and redundant vision tokens is essential for accurate error minimization.
    \vspace{-2mm}
\end{itemize}

\begin{table}[t]
\centering
\caption{Ablation study of Affinity-Aware Hessian (VEQ-MA) on downstream tasks. $p$: Router confidence; $\alpha$: Modality indicator. }
\label{tab:ablation_veq2}
\vspace{-1mm}
\resizebox{\columnwidth}{!}{
\begin{tabular}{lcccc}
\toprule
\rowcolor{colorhead}
Config & MMMU & InfoVQA & ScienceQA & Avg. \\
\midrule
w/o $p$ ($p=1$) & 42.33 & 64.30 & 83.35 & 63.63 \\
w/o $\alpha$ ($\alpha=1$) & 41.33 & 63.72 & 82.34 & 62.46 \\
\textbf{VEQ-MA (Full)} & \textbf{42.56} & \textbf{64.48} & \textbf{84.85} & \textbf{63.96} \\
\bottomrule
\end{tabular}
}
\vspace{-6mm}
\end{table}

\noindent\textbf{Parameter Sensitivity Analysis.}
To further validate the robustness of our method, we analyze the average Perplexity (PPL) on 64 samples under varying parameter configurations. These samples are randomly extracted from MMMU \cite{yue_mmmu_2024} validation dataset. We perform a grid search for the variable pairs in VEQ-ME ($\gamma, \beta$) and VEQ-MA ($\lambda, \gamma$).  It is worth noting that while the formulation of VEQ-MA is theoretically governed by the affinity $p$ and sensitivity ratio $\alpha$, directly tuning $p$ and $\alpha$ lacks intuitive interpretability. Consequently, we choose to adopt ($\lambda, \gamma$) as the variable pair for this ablation study. 

\noindent\textit{1) Sensitivity of VEQ-ME ($\gamma$ vs. $\beta$).}
In our formulation, $\gamma$ represents the sensitivity of text tokens relative to visual tokens, while $\beta$ accounts for the quantity ratio between the two modalities. These hyperparameters determine the expert importance, which guides the search for the optimal quantization parameters.
As illustrated in Figure \ref{fig:ablation_heatmap}, we observe that proportionally scaling $\gamma$ and $\beta$ yields consistent PPL values. This implies that the search for quantization parameters depends on the \textit{relative} ratio of expert importance.

\noindent\textit{2) Sensitivity of VEQ-MA ($\lambda$ vs. $\gamma$).}
We analyze the Hessian weighting by varying the modality sensitivity ratio $\alpha$ and the affinity strength $p$. Specifically, we treat the variation of $p$ using a modulation coefficient $\lambda$, which controls the intensity of router confidence. We define the effective affinity as a weighted interpolation: $(1-\lambda) \cdot \mathbf{1} + \lambda \cdot \mathbf{p}$. Under this formulation, $\lambda=1$ represents the raw router confidence, while $\lambda=0$ indicates uniform affinity.As shown in Figure \ref{fig:ablation_heatmap}, the model achieves optimal stability (lowest PPL of 2.2526) at the full configuration ($\lambda = 1, \gamma=\gamma_0$). 

\vspace{-2mm}
\section{Conclusion} 
\vspace{-1mm}
In this work, we presented \textbf{V}isual \textbf{E}xpert \textbf{Q}uantization (\textbf{VEQ}), a specialized post-training quantization framework designed to address the unique challenges of compressing Mixture-of-Experts Vision-Language Models (MoE VLMs). By transcending the limitations of treating MoE FFNs as monolithic dense structures, VEQ effectively addresses both the inherent sparsity of expert activations and the statistical heterogeneity between visual and textual modalities. Across a diverse set of multimodal benchmarks, VEQ consistently outperforms established baselines. By aligning quantization strategies with the structural and modal properties of MoE VLMs, VEQ establishes a new state-of-the-art, paving the way for the efficient deployment of large-scale multimodal agents in resource-constrained environments.

\bibliography{arxiv}

@article{openai_gpt-4o_2024,
  title={Gpt-4o system card},
  author={Hurst, Aaron and Lerer, Adam and Goucher, Adam P and Perelman, Adam and Ramesh, Aditya and Clark, Aidan and Ostrow, AJ and Welihinda, Akila and Hayes, Alan and Radford, Alec and others},
  journal={arXiv preprint arXiv:2410.21276},
  year={2024}
}

@article{deepseek-vl2,
    title={DeepSeek-VL2: Mixture-of-Experts Vision-Language Models for Advanced Multimodal Understanding},
    author={Wu, Zhiyu and Chen, Xiaokang and Pan, Zizheng and Liu, Xingchao and Liu, Wen and Dai, Damai and Gao, Huazuo and Ma, Yiyang and Wu, Chengyue and Wang, Bingxuan and others},
    journal={arXiv preprint arXiv:2412.10302},
    year={2024},
}

@article{team2025kimi,
  title={Kimi-vl technical report},
  author={Team, Kimi and Du, Angang and Yin, Bohong and Xing, Bowei and Qu, Bowen and Wang, Bowen and Chen, Cheng and Zhang, Chenlin and Du, Chenzhuang and Wei, Chu and others},
  journal={arXiv preprint arXiv:2504.07491},
  year={2025}
}

@article{Qwen3-VL,
  title={Qwen3-VL Technical Report}, 
  author={Shuai Bai and Yuxuan Cai and Ruizhe Chen and Keqin Chen and Xionghui Chen and Zesen Cheng and Lianghao Deng and Wei Ding and Chang Gao and Chunjiang Ge and Wenbin Ge and Zhifang Guo and others},
  journal={arXiv preprint arXiv:2511.21631},
  year={2025}
}

@misc{ernie2025technicalreport,
      title={ERNIE 4.5 Technical Report},
      author={Baidu-ERNIE-Team},
      year={2025},
      eprint={},
      archivePrefix={arXiv},
      primaryClass={cs.CL},
      url={}
}

@inproceedings{radford_learning_2021,
  title={Learning transferable visual models from natural language supervision},
  author={Radford, Alec and Kim, Jong Wook and Hallacy, Chris and Ramesh, Aditya and Goh, Gabriel and Agarwal, Sandhini and Sastry, Girish and Askell, Amanda and Mishkin, Pamela and Clark, Jack and others},
  booktitle={ICML},
  year={2021}
}

@inproceedings{alayrac_flamingo_2022,
  title={Flamingo: a visual language model for few-shot learning},
  author={Alayrac, Jean-Baptiste and Donahue, Jeff and Luc, Pauline and Miech, Antoine and Barr, Iain and Hasson, Yana and Lenc, Karel and Mensch, Arthur and Millican, Katherine and Reynolds, Malcolm and others},
  booktitle={NeurIPS},
  year={2022}
}

@inproceedings{li_blip-2_2023,
  title={Blip-2: Bootstrapping language-image pre-training with frozen image encoders and large language models},
  author={Li, Junnan and Li, Dongxu and Savarese, Silvio and Hoi, Steven},
  booktitle={ICML},
  year={2023}
}

@inproceedings{lu_vilbert_2019,
  title={Vilbert: Pretraining task-agnostic visiolinguistic representations for vision-and-language tasks},
  author={Lu, Jiasen and Batra, Dhruv and Parikh, Devi and Lee, Stefan},
  booktitle={NeurIPS},
  year={2019}
}

@inproceedings{lin_awq_2024,
  title={Awq: Activation-aware weight quantization for on-device llm compression and acceleration},
  author={Lin, Ji and Tang, Jiaming and Tang, Haotian and Yang, Shang and Chen, Wei-Ming and Wang, Wei-Chen and Xiao, Guangxuan and Dang, Xingyu and Gan, Chuang and Han, Song},
  booktitle={MLSys},
  year={2024}
}

@article{fedus_switch_2022,
  title={Switch transformers: Scaling to trillion parameter models with simple and efficient sparsity},
  author={Fedus, William and Zoph, Barret and Shazeer, Noam},
  journal={JMLR},
  year={2022}
}

@article{frantar_gptq_2023,
  title={Gptq: Accurate post-training quantization for generative pre-trained transformers},
  author={Frantar, Elias and Ashkboos, Saleh and Hoefler, Torsten and Alistarh, Dan},
  journal={arXiv preprint arXiv:2210.17323},
  year={2022}
}

@inproceedings{xiao_smoothquant_2024,
  title={Smoothquant: Accurate and efficient post-training quantization for large language models},
  author={Xiao, Guangxuan and Lin, Ji and Seznec, Mickael and Wu, Hao and Demouth, Julien and Han, Song},
  booktitle={ICML},
  year={2024}
}

@article{liu_spinquant_2025,
  title={Spinquant: Llm quantization with learned rotations},
  author={Liu, Zechun and Zhao, Changsheng and Fedorov, Igor and Soran, Bilge and Choudhary, Dhruv and Krishnamoorthi, Raghuraman and Chandra, Vikas and Tian, Yuandong and Blankevoort, Tijmen},
  journal={arXiv preprint arXiv:2405.16406},
  year={2024}
}

@inproceedings{chi_representation_2022,
  title={On the representation collapse of sparse mixture of experts},
  author={Chi, Zewen and Dong, Li and Huang, Shaohan and Dai, Damai and Ma, Shuming and Patra, Barun and Singhal, Saksham and Bajaj, Payal and Song, Xia and Mao, Xian-Ling and others},
  booktitle={NeurIPS},
  year={2022}
}

@inproceedings{liang_mind_2022,
  title={Mind the gap: Understanding the modality gap in multi-modal contrastive representation learning},
  author={Liang, Victor Weixin and Zhang, Yuhui and Kwon, Yongchan and Yeung, Serena and Zou, James Y},
  booktitle={NeurIPS},
  year={2022}
}

@inproceedings{yue_mmmu_2024,
  title={Mmmu: A massive multi-discipline multimodal understanding and reasoning benchmark for expert agi},
  author={Yue, Xiang and Ni, Yuansheng and Zhang, Kai and Zheng, Tianyu and Liu, Ruoqi and Zhang, Ge and Stevens, Samuel and Jiang, Dongfu and Ren, Weiming and Sun, Yuxuan and others},
  booktitle={CVPR},
  year={2024}
}

@inproceedings{kembhavi_diagram_2016,
  title={A diagram is worth a dozen images},
  author={Kembhavi, Aniruddha and Salvato, Mike and Kolve, Eric and Seo, Minjoon and Hajishirzi, Hannaneh and Farhadi, Ali},
  booktitle={ECCV},
  year={2016}
}

@inproceedings{singh2019towards,
  title={Towards vqa models that can read},
  author={Singh, Amanpreet and Natarajan, Vivek and Shah, Meet and Jiang, Yu and Chen, Xinlei and Batra, Dhruv and Parikh, Devi and Rohrbach, Marcus},
  booktitle={CVPR},
  year={2019}
}

@inproceedings{mathew_infographicvqa_2021,
  title={Infographicvqa},
  author={Mathew, Minesh and Bagal, Viraj and Tito, Rub{\`e}n and Karatzas, Dimosthenis and Valveny, Ernest and Jawahar, CV},
  booktitle={WACV},
  year={2022}
}

@article{xue2025vlmq,
  title={Vlmq: Efficient post-training quantization for large vision-language models via hessian augmentation},
  author={Xue, Yufei and Huang, Yushi and Shao, Jiawei and Zhang, Jun},
  journal={arXiv preprint arXiv:2508.03351},
  year={2025}
}

@inproceedings{wang2024q-vlm,
  title={Q-vlm: Post-training quantization for large vision-language models},
  author={Wang, Changyuan and Wang, Ziwei and Xu, Xiuwei and Tang, Yansong and Zhou, Jie and Lu, Jiwen},
  booktitle={NeurIPS},
  year={2024}
}

@inproceedings{li2025mbq,
  title={Mbq: Modality-balanced quantization for large vision-language models},
  author={Li, Shiyao and Hu, Yingchun and Ning, Xuefei and Liu, Xihui and Hong, Ke and Jia, Xiaotao and Li, Xiuhong and Yan, Yaqi and Ran, Pei and Dai, Guohao and others},
  booktitle={CVPR},
  year={2025}
}

@article{wang2025bi-vlm,
  title={Bi-VLM: Pushing Ultra-Low Precision Post-Training Quantization Boundaries in Vision-Language Models},
  author={Wang, Xijun and Huang, Junyun and Abdalla, Rayyan and Zhang, Chengyuan and Xian, Ruiqi and Manocha, Dinesh},
  journal={arXiv preprint arXiv:2509.18763},
  year={2025}
}

@article{yu2025mquant,
  title={Mquant: Unleashing the inference potential of multimodal large language models via full static quantization},
  author={Yu, JiangYong and Zhou, Sifan and Yang, Dawei and Wang, Shuo and Li, Shuoyu and Hu, Xing and Xu, Chen and Xu, Zukang and Shu, Changyong and Yuan, Zhihang},
  journal={arXiv preprint arXiv:2502.00425},
  year={2025}
}

@article{hu2025moequant,
  title={MoEQuant: Enhancing Quantization for Mixture-of-Experts Large Language Models via Expert-Balanced Sampling and Affinity Guidance},
  author={Hu, Xing and Chen, Zhixuan and Yang, Dawei and Xu, Zukang and Xu, Chen and Yuan, Zhihang and Zhou, Sifan and Yu, Jiangyong},
  journal={arXiv preprint arXiv:2505.03804},
  year={2025}
}

@article{zheng2025moqa,
  title={MoQa: Rethinking MoE Quantization with Multi-stage Data-model Distribution Awareness},
  author={Zheng, Zihao and Cui, Xiuping and Zheng, Size and Li, Maoliang and Chen, Jiayu and Liang, Yun and Chen, Xiang},
  journal={arXiv preprint arXiv:2503.21135},
  year={2025}
}

@article{zhang2025moqe,
  title={MoQE: Improve Quantization Model performance via Mixture of Quantization Experts},
  author={Zhang, Jinhao and Zhang, Yunquan and Zhang, Boyang and Liu, Zeyu and Cheng, Daning},
  journal={arXiv preprint arXiv:2508.09204},
  year={2025}
}

@article{duanmu2025mxmoe,
  title={MxMoE: Mixed-precision Quantization for MoE with Accuracy and Performance Co-Design},
  author={Duanmu, Haojie and Li, Xiuhong and Yuan, Zhihang and Zheng, Size and Duan, Jiangfei and Zhang, Xingcheng and Lin, Dahua},
  journal={arXiv preprint arXiv:2505.05799},
  year={2025}
}

@dataset{realworldqa,
  title       = {RealWorldQA: A Real-World Multimodal Question Answering Benchmark},
  author      = {Team, xAI},
  year        = {2024},

  url         = {https://huggingface.co/datasets/xai-org/RealworldQA}
}

@inproceedings{lu2022learn,
  title={Learn to explain: Multimodal reasoning via thought chains for science question answering},
  author={Lu, Pan and Mishra, Swaroop and Xia, Tanglin and Qiu, Liang and Chang, Kai-Wei and Zhu, Song-Chun and Tafjord, Oyvind and Clark, Peter and Kalyan, Ashwin},
  booktitle={NeurIPS},
  year={2022}
}

@inproceedings{gurari2019vizwiz,
  title={Vizwiz-priv: A dataset for recognizing the presence and purpose of private visual information in images taken by blind people},
  author={Gurari, Danna and Li, Qing and Lin, Chi and Zhao, Yinan and Guo, Anhong and Stangl, Abigale and Bigham, Jeffrey P},
  booktitle={CVPR},
  year={2019}
}

@article{lmms_eval2024,
    title={LMMs-Eval: Reality Check on the Evaluation of Large Multimodal Models}, 
    author={Kaichen Zhang and Bo Li and Peiyuan Zhang and Fanyi Pu and Joshua Adrian Cahyono and Kairui Hu and Shuai Liu and Yuanhan Zhang and Jingkang Yang and Chunyuan Li and Ziwei Liu},
    journal={arXiv preprint arXiv:2407.12772},   
    year={2024},
}

@inproceedings{zheng2024sglang,
  title={Sglang: Efficient execution of structured language model programs},
  author={Zheng, Lianmin and Yin, Liangsheng and Xie, Zhiqiang and Sun, Chuyue Livia and Huang, Jeff and Yu, Cody Hao and Cao, Shiyi and Kozyrakis, Christos and Stoica, Ion and Gonzalez, Joseph E and others},
  booktitle={NeurIPS},
  year={2024}
}

@inproceedings{lin2014coco,
  title={Microsoft coco: Common objects in context},
  author={Lin, Tsung-Yi and Maire, Michael and Belongie, Serge and Hays, James and Perona, Pietro and Ramanan, Deva and Doll{\'a}r, Piotr and Zitnick, C Lawrence},
  booktitle={ECCV},
  year={2014}
}

@article{zhang2024mme,
  title={Mme-realworld: Could your multimodal llm challenge high-resolution real-world scenarios that are difficult for humans?},
  author={Zhang, Yi-Fan and Zhang, Huanyu and Tian, Haochen and Fu, Chaoyou and Zhang, Shuangqing and Wu, Junfei and Li, Feng and Wang, Kun and Wen, Qingsong and Zhang, Zhang and others},
  journal={arXiv preprint arXiv:2408.13257},
  year={2024}
}

@inproceedings{liu2024mmbench,
  title={Mmbench: Is your multi-modal model an all-around player?},
  author={Liu, Yuan and Duan, Haodong and Zhang, Yuanhan and Li, Bo and Zhang, Songyang and Zhao, Wangbo and Yuan, Yike and Wang, Jiaqi and He, Conghui and Liu, Ziwei and others},
  booktitle={ECCV},
  year={2024}
}
\bibliographystyle{icml2026}

\end{document}